\documentclass{article} 
\PassOptionsToPackage{table}{xcolor}
\usepackage{iclr2027_conference,times}

\usepackage{amsmath,amsfonts,bm}

\def\eqref#1{equation~\ref{#1}}

\def\1{\bm{1}}

\DeclareMathAlphabet{\mathsfit}{\encodingdefault}{\sfdefault}{m}{sl}
\SetMathAlphabet{\mathsfit}{bold}{\encodingdefault}{\sfdefault}{bx}{n}

\usepackage{times}
\usepackage{pifont}
\usepackage{graphicx}
\usepackage{booktabs}
\usepackage{caption}
\usepackage{pifont}
\usepackage{hyperref}
\usepackage{url}
\usepackage{bm}
\usepackage{multirow}
\usepackage{amsmath}
\usepackage{amssymb}
\usepackage[table]{xcolor}
\usepackage{listings}
\usepackage{subcaption}
\usepackage{wrapfig}
\usepackage{tikz}
\usepackage{algorithm}
\usepackage{algpseudocode}
\usepackage{caption}
\usepackage{float}

\usetikzlibrary{shapes,arrows,positioning,fit,calc}

\newcommand{\ironmanicon}{%
  \includegraphics[height=2.0em]{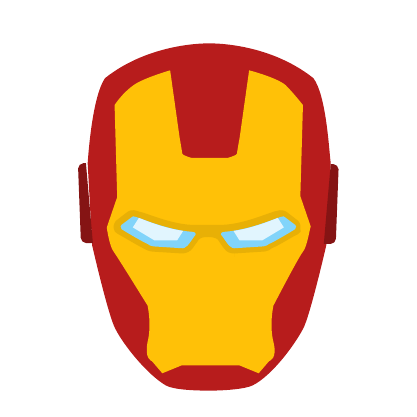}%
}

\definecolor{lightbrown}{RGB}{245,230,200}
\definecolor{goodgreen}{RGB}{0,128,0}
\definecolor{badred}{RGB}{200,0,0}
\title{Sometimes You Gotta Run Before You Can Walk: Run-then-Walk Scheduling Strategy for VLM Autonomous Driving}

\author{
\textbf{Yuqi Ye\textsuperscript{1}, Shangkun Sun\textsuperscript{1}, Junhong Lin\textsuperscript{1}, Jiayi Zhao\textsuperscript{1}, Changhao Peng\textsuperscript{1},} \\
\hspace*{0.2em}\textbf{Wei Zheng\textsuperscript{2}, Guoqing Liu\textsuperscript{2}, Tiesong Zhao\textsuperscript{3}, Wei Gao\textsuperscript{1,$^\dagger$}} \\
\textsuperscript{1}Peking University, \textsuperscript{2}Minieye Technology Co., Ltd, \textsuperscript{3}Fuzhou University \\
\vspace{-1.0cm}
}

\iclrfinalcopy 
\begin{document}

\maketitle

\newcommand\blfootnote[1]{%
\begingroup
\renewcommand\thefootnote{}%
\hypersetup{linkbordercolor=white,pdfborder={0 0 0}}
\footnote{#1}%
\addtocounter{footnote}{-1}%
\endgroup
}

\vspace{-0.3cm}
\begin{center}
    \ironmanicon 
    \hspace{0.4em}%
    \begin{tabular}[c]{@{}l@{}}
        \emph{``Sometimes You Gotta Run Before You Can Walk.''}\\
        \multicolumn{1}{c@{}}{\textemdash\ \textit{Iron Man} (2008)}
    \end{tabular}
\end{center}

\begin{figure*}[h]
\centering
\vspace{-0.5cm}
\includegraphics[width=\linewidth]{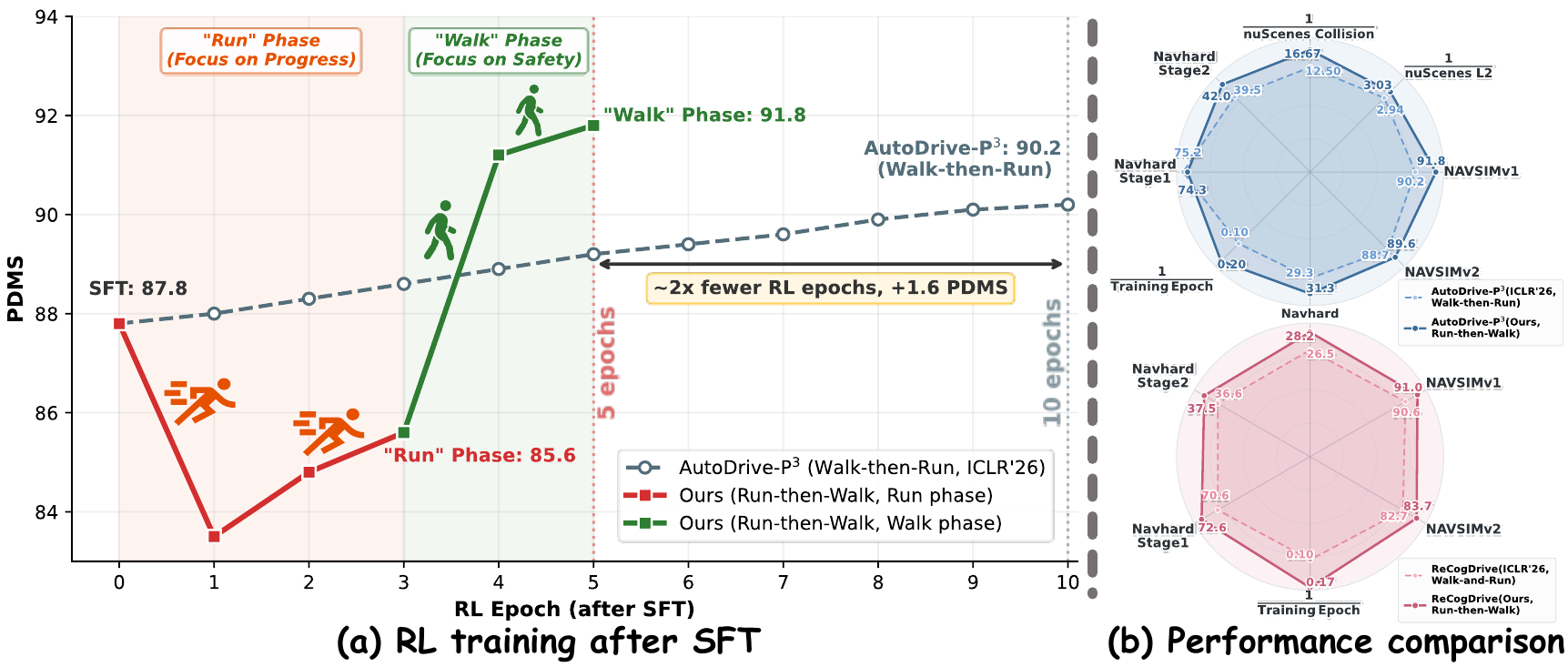}
\vspace{-0.7cm}
\caption{\textbf{RL training and performance comparison.} 
(a) Our \textit{Run-then-Walk} strategy reaches a higher PDMS of 91.8 in only 5 RL epochs with AutoDrive-P$^3$ planner~\citep{ye2026autodrive}, outperforming 90.2 PDMS achieved by the 10-epoch baseline while substantially accelerating convergence.
(b) Radar-chart comparisons of various methods on different benchmarks. With \textit{Run-then-Walk}, the evaluated planner families improve their aggregate scores on the reported datasets while training converges 40--50\% faster.
}
\label{fig:convergence_comparison}
\end{figure*}

\begin{abstract}

Recent VLM-based autonomous driving planners adopt GRPO-style reinforcement learning to optimize driving performance. However, existing GRPO recipes either optimize driving efficiency, risking progress-seeking but unsafe behavior, or enforce early safety constraints, leading to overly conservative behavior; both require lengthy training. To solve these problems, we first reveal two distinct RL regimes: a progress regime (Run-GRPO) that aggressively explores high progress, and a safety regime (Walk-GRPO) that restores safety under stable progress. Based on this finding, we propose $\textit{Run-then-Walk}$, a simple yet effective two-stage reward scheduling strategy for GRPO, achieving both better performance and faster convergence. Unlike one-stage RL, which may focus on progress, safety, or a mixture of both within a single training phase, this schedule explicitly separates progress discovery from safety repair. In the $\textit{Run}$ phase, we focus on progress, allowing the policy to escape the conservative bias and discover high-progress modes. In the subsequent $\textit{Walk}$ phase, we introduce endpoint and safety strategy to repair unsafe behaviors from the Run phase. This reversed schedule overcomes the conservatism of Walk-first methods and the unsafe progress-seeking of joint optimization. We validate it with various VLM-based planners on multiple benchmarks: NAVSIMv1, NAVSIMv2, Navhard, and nuScenes. Extensive experiments demonstrate improved driving performance while requiring 40--50\% fewer RL training epochs than the baselines. Code is available at \url{https://github.com/haha-yuki-haha/AutoDrive-P3_with_Run-then-walk}.
\end{abstract}

\section{Introduction}
\label{sec:intro}

End-to-end autonomous driving has evolved from compact architectures~\citep{hu2023planning, hu2022st, jiang2023vad, liao2025diffusiondrive, li2025end, li2024enhancing} to Vision-Language Model (VLM) based systems~\citep{tian2024drivevlm, hwang2024emma, xu2024vlm, xing2025openemma}, as they leverage large-scale pre-training with world knowledge. While VLMs achieve strong results via SFT, recent VLM planners adopt GRPO-style RL~\citep{shao2024deepseekmath, guo2025deepseek} to achieve better performance, especially in closed-loop driving quality~\citep{dauner2024navsim, cao2025pseudo}. Composite closed-loop metrics entangle progress and safety, and balancing these conflicting objectives remains the central challenge of RL for autonomous driving.

\begin{figure*}[t]
\centering
\includegraphics[width=0.9\textwidth]{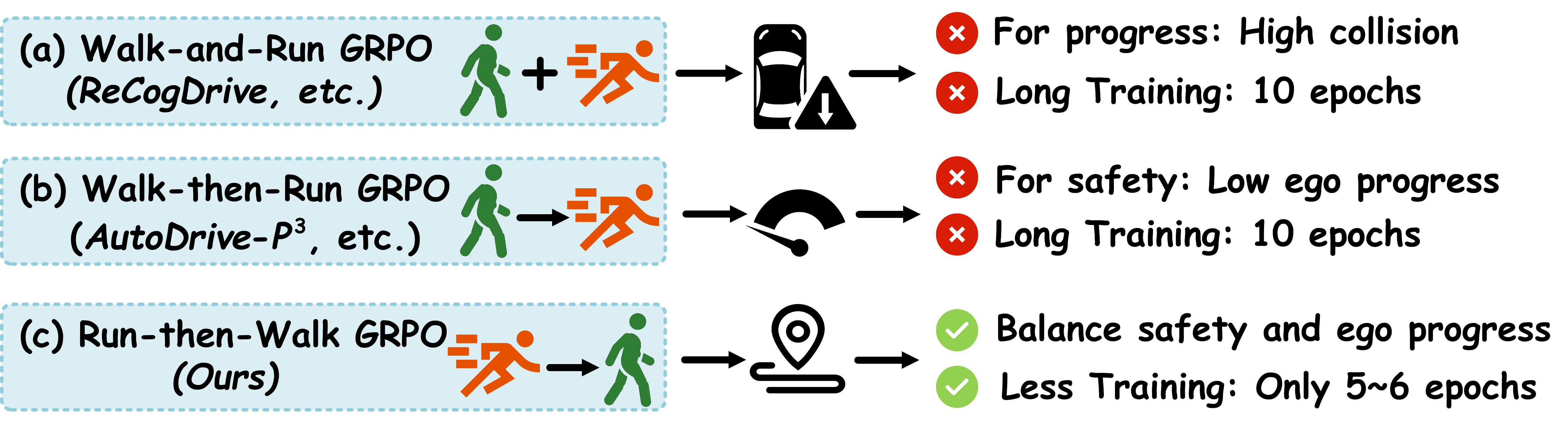}
\caption{\textbf{Comparison of GRPO training paradigms.} (a) The \textit{Walk-and-Run} paradigm, like ReCogDrive \citep{li2025recogdrive}, remains biased toward Run, resulting in high collision. (b) The \textit{Walk-then-Run} paradigm, seen in AutoDrive-P$^3$ \citep{ye2026autodrive}, prioritizes safety at the cost of low ego progress. (c) Our proposed \textit{Run-then-Walk} paradigm first explores freely and then refines for safety, achieving a balance between safety and progress while saving 40--50\% of the training steps.}
\label{fig:intro_paradigm}
\vspace{-0.25cm}
\end{figure*}

To address this issue, more structured training strategies have been explored~\citep{zhou2025autovla, jiang2025alphadrive, li2025recogdrive, ye2026autodrive}. However, as illustrated in Fig.~\ref{fig:intro_paradigm}, existing RL strategies still suffer from three major limitations. \textbf{1) Walk-and-Run: High collision.} ReCogDrive~\citep{li2025recogdrive} uses Walk-and-Run style objectives, but its training remains biased toward progress. The resulting policy achieves progress with a high collision rate and other safety violations. \textbf{2) Walk-then-Run: Over-conservatism.} AutoDrive-P$^3$~\citep{ye2026autodrive} starts from strict safety constraints and only gradually relaxes them, making the model overly conservative and hesitant to move forward (Fig.~\ref{fig:intro_paradigm}(b)). \textbf{3) Lengthy training.} Both baselines require 10 RL epochs, which is extremely time-consuming and consumes substantial GPU resources. This naturally raises a question: \textit{can we find a faster and more effective RL training paradigm for VLM-based driving?}

To answer this question, we first conduct a controlled toy experiment using AutoDrive-P$^3$ on NAVSIM as a concrete training instance. Starting from the same SFT checkpoint, we compare direct Run-GRPO (progress-oriented) and direct Walk-GRPO (safety-constrained) as the first RL objective. As shown in Fig.~\ref{fig:toy_run_walk}, direct Run-GRPO first rapidly optimizes ego progress (EP), and only after EP saturates does the policy begin to repair safety metrics; however, without explicit safety constraints, improvement within this aggressive regime is limited. Direct Walk-GRPO exhibits the opposite pattern: the policy explores only within the safe region around the SFT solution, preserving safety at the cost of slow progress improvement. \textbf{\textit{Our key finding is that RL exhibits two distinct regimes: a progress regime (Run-GRPO) that aggressively explores high progress and a safety regime (Walk-GRPO) that keeps safe under low progress.}}

Motivated by this diagnosis, we propose \textit{Run-then-Walk}, a simple yet effective two-stage scheduling strategy for GRPO. Unlike one-stage RL strategy, which may focus on progress, safety, or a mixture of both within a single training phase, our strategy explicitly separates progress in an ordered manner: the \textit{Run} phase first encourages progress exploration under relaxed constraints, and the \textit{Walk} phase then introduces endpoint and safety objectives to repair unsafe behaviors. As illustrated in Fig.~\ref{fig:intro_paradigm}(c), this schedule differs both from one-stage RL and from generic easy-to-hard curricula, while addressing the unsafe behavior of \textit{Walk-and-Run} and the over-conservatism of \textit{Walk-then-Run}.

As shown in Fig.~\ref{fig:convergence_comparison}, this algorithm yields both faster convergence and higher final performance. Across different VLM-based planners and multiple benchmarks, it improves driving performance while requiring only 5 epochs for AutoDrive-P$^3$ and 6 epochs for ReCogDrive, fewer than the 10 epochs used by the original RL policies. The final policies retain high EP and restore collision, drivable-area, and time-to-collision safety metrics. We hope that this progress-first, safety-refinement paradigm, simple and effective, can extend beyond autonomous driving to broader VLA decision-making problems. The main contributions of this paper are summarized as follows:
\begin{itemize}
    \item We reveal that the RL optimization direction for VLM-based driving critically affects the policy, and summarize it into two main regimes: a progress regime (Run-GRPO) that aggressively explores high progress at the cost of safety, and a safety regime (Walk-GRPO) that preserves safety but remains conservative with slow progress improvement.
    \item We introduce \textit{Run-then-Walk}, a simple and effective two-stage RL training strategy that first expands the progress distribution under relaxed constraints and then repairs safety with endpoint and safety-related objectives, which differs both from one-stage mixed-objective RL and from generic easy-to-hard curricula.
    \item We validate the same algorithm on VLM-based autoregressive and diffusion planners across NAVSIMv1, NAVSIMv2, Navhard, and nuScenes. The results show higher driving performance, and 40--50\% fewer RL training steps.
\end{itemize}

\section{Related Work}
\label{sec:related}
\vspace{-0.25cm}

\textbf{VLM-based end-to-end driving.}
End-to-end driving has evolved from modular pipelines such as UniAD~\citep{hu2023planning}, VAD~\citep{jiang2023vad}, Law~\citep{li2024enhancing}, Wote~\citep{li2025end}, and DiffusionDrive~\citep{liao2025diffusiondrive} to VLM-based systems including DriveVLM~\citep{tian2024drivevlm}, EMMA~\citep{hwang2024emma}, VLM-AD~\citep{xu2024vlm}, and OpenEMMA~\citep{xing2025openemma}. These VLM planners either use VLM features to support a downstream trajectory generator, as in ReCogDrive~\citep{li2025recogdrive}, or directly emit trajectories as language tokens, like OpenDriveVLA~\citep{zhou2025opendrivevla}, AutoVLA~\citep{zhou2025autovla}, and AutoDrive-P$^3$~\citep{ye2026autodrive}. To validate the robustness of our method, we evaluate it on two representative VLM planners: the autoregressive AutoDrive-P$^3$ and the diffusion-based ReCogDrive.

\textbf{Group Relative Policy Optimization.}
Group Relative Policy Optimization (GRPO)~\citep{shao2024deepseekmath, guo2025deepseek, shao2025deepseekmath}, originally proposed by DeepSeek, has emerged as an effective reinforcement learning algorithm for enhancing the reasoning capabilities of large language models. Its success has since been extended to vision-language models, with works such as Vision-R1~\citep{huang2025vision} demonstrating that GRPO can substantially improve VLM reasoning. In the autonomous driving domain, several recent efforts have adopted GRPO to fine-tune driving VLMs with closed-loop metrics as rewards. However, existing approaches lie between two extremes: mixed Walk-and-Run schedules that remain biased toward progress~\citep{li2025recogdrive}, which may preserve progress-seeking but unsafe behaviors, and safety-first schedules that relax constraints only gradually~\citep{ye2026autodrive}, which may suppress progress and require a long training horizon. These limitations indicate that how different RL training regimes shape the progress-safety trade-off in VLA-based planners remains poorly understood. Systematically characterizing these regimes is therefore essential for designing RL strategies with better performance and faster convergence.

\section{Toy Experiment: Run-GRPO vs. Walk-GRPO from SFT}
\label{sec:toy_exp}
\begin{figure*}[t]
\centering
\includegraphics[width=0.95\textwidth]{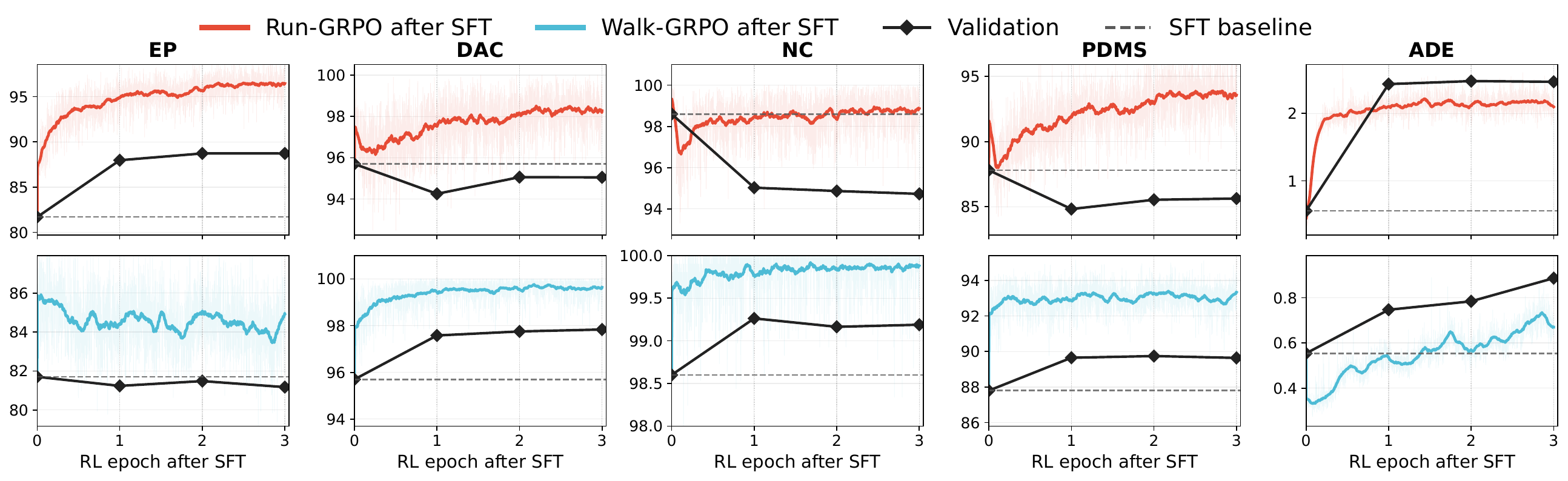}
\vspace{-0.3cm}
\caption{\textbf{Toy experiment from the same SFT checkpoint.} Starting from the same SFT model, we compare direct Run GRPO and direct constrained Walk GRPO for three epochs. Direct Run quickly discovers aggressive high-progress behaviors, but then has to improve safety within this overly aggressive regime, causing PDMS to fall well below the SFT baseline. Direct Walk exhibits the opposite pattern: it explores only within a conservative safety regime, so PDMS improves slowly.}
\label{fig:toy_run_walk}
\vspace{-0.5cm}
\end{figure*}

We conduct a toy experiment to understand how the first RL objective shapes policy learning after SFT. Using the AutoDrive-P$^3$ planner in NAVSIMv1 benchmark as an illustrative example, we start from its SFT checkpoint, which already exhibits low-ADE yet relatively low-progress behavior, and compare two direct RL fine-tuning strategies. In direct Run-GRPO, we use a PDMS metric dominated by ego progress as the planning reward. Although PDMS is a composite reward, EP accounts for \(5/12\) of its total weight, i.e., nearly half, and EP is also a relatively easier objective to optimize, so the policy tends to learn high-progress behaviors. In direct Walk-GRPO, we instead use a constrained planning reward consisting of the endpoint reward together with the safety terms DAC and NC from PDMS. All other settings are kept the same, and Fig.~\ref{fig:toy_run_walk} reports the resulting EP, DAC, NC, PDMS, and ADE trends during training.

Fig.~\ref{fig:toy_run_walk} shows that direct Run-GRPO and direct Walk-GRPO fail in opposite ways. With direct Run-GRPO, the policy quickly discovers that ego progress (EP) is the easiest part of PDMS to optimize from the SFT initialization. It therefore learns aggressive forward-driving behaviors first, which raises EP but also increases collisions and drivable-area violations. As a consequence, the overall PDMS drops below the SFT baseline even though progress improves. Only after the policy has already entered this aggressive regime does optimization begin to repair safety. At that point, however, the model can only search for trajectories that are as safe as possible under aggressive behavior, making it difficult to recover a truly balanced policy.

Direct Walk-GRPO exhibits the opposite pattern. Because its planning reward is dominated from the beginning by endpoint consistency and the safety terms DAC and NC, the policy is encouraged to explore only within a constrained safe region around the SFT solution. This preserves and improves safety-related behavior, but it also suppresses forward exploration and keeps the policy conservative. As a result, PDMS improves only slowly, since the model learns to drive safely under constraints rather than to discover stronger progress modes.

This contrast directly motivates our Run-then-Walk strategy. A good first-stage objective after SFT should be able to break the conservative bias and uncover progress-oriented behaviors, while a good second-stage objective should repair the safety deficits introduced by that exploration. Run-then-Walk follows exactly this logic: it first uses Run to expand the progress distribution, and then uses Walk to refine the policy into safe and stable closed-loop driving.

\section{Method}
\label{sec:method}

\subsection{Run-then-Walk Planning Strategy Design}
Let $\pi_\theta$ be a VLM driving policy, for a driving query $q$, a complete policy sample $x_i$ can be an autoregressive response or a continuous sample from a diffusion trajectory head; a decoder maps either form to a trajectory $\tau_i$. Run-then-Walk preserves the planner's architecture and native auxiliary objectives and changes only temporal schedule and the planning reward on the decoded trajectory distribution rather than on a particular output representation.

In the Run stage, we adopt an aggressive planning reward that prioritizes ego progress,
\begin{equation}
R_{\mathrm{plan}}^{\mathrm{run}}(\tau_i)=PDMS(\tau_i),
\end{equation}
where $PDMS(\tau_i)$ denotes an aggressive reward function. This stage intentionally relaxes safety repair and encourages the policy to move beyond the conservative SFT solution by discovering higher-progress trajectory modes. 

In the Walk stage, we keep the other reward terms unchanged and replace the aggressive planning reward with a new planning objective composed of endpoint and safety-related reward terms. We design the endpoint distance $d_E$ as the $L_1$ distance between the final predicted point $p_T$ and the expert endpoint $g_T$:
\begin{equation}
d_E = \|p_T - g_T\|_1.
\end{equation}
We introduce a step size hyperparameter $\Delta>0$ and define the endpoint reward as
\begin{equation}
k(d_E)=
\min\!\left(
K,\;
\max\!\left(
0,\left\lfloor (d_E-2\Delta)/\Delta\right\rfloor + 1
\right)
\right),
\qquad
R_{\mathrm{end}}(d_E)=1-\eta\,k(d_E),
\label{eq:endpoint_reward}
\end{equation}
where $\eta \in (0,1)$ is the reward decrement and $K=\lfloor 1/\eta \rfloor$ is the maximum number of decay steps. This yields a piecewise endpoint reward that decreases by a fixed amount every $\Delta$ units of endpoint error and eventually becomes zero. A smaller $\Delta$ enforces a tighter endpoint constraint, making the policy more conservative, while a larger $\Delta$ yields a looser Walk stage, making it more aggressive. Additionally, a larger $\eta$ causes the endpoint reward to decay more aggressively. With the endpoint reward, the Walk planning reward is
\begin{equation}
R_{\mathrm{plan}}^{\mathrm{walk}}(\tau_i,\tau_i^\star)
=
\mathbb{I}_{\mathrm{safe}}(\tau_i)\,
\Big(R_{\mathrm{safe}}(\tau_i)+R_{\mathrm{end}}(d_E)\Big),
\label{eq:walk_plan_reward}
\end{equation}
where $\mathbb{I}_{\mathrm{safe}}(\tau_i)=1$ only if the rollout satisfies all safety constraints, and is $0$ otherwise. Here $\tau_i^\star$ denotes the expert trajectory, and $R_{\mathrm{safe}}(\cdot)$ denotes a safety-related reward term. Therefore, once a rollout becomes unsafe, its planning reward is set to zero; otherwise, the Walk reward combines endpoint consistency with safety-related terms. 
In our implementation, we instantiate $R_{\mathrm{safe}}(\tau_i)$ as $\mathrm{NC}(\tau_i)+\mathrm{DAC}(\tau_i)$, and define $\mathbb{I}_{\mathrm{safe}}(\tau_i)=1$ only if $\mathrm{NC}(\tau_i)>0$,  and $\mathrm{DAC}(\tau_i)>0$.

Because both stages consume only a decoded trajectory and closed-loop feedback, the construction is planner-agnostic. For an autoregressive planner, the reward is assigned to the sampled token sequence that contains the trajectory; for a VLM-conditioned diffusion planner, it is assigned to the sampled continuous trajectory. In both cases, Run expands high-progress modes and Walk applies the same endpoint term and safety gate while preserving the planner's native auxiliary supervision.

\begin{figure*}[t]
\centering
\includegraphics[width=\textwidth]{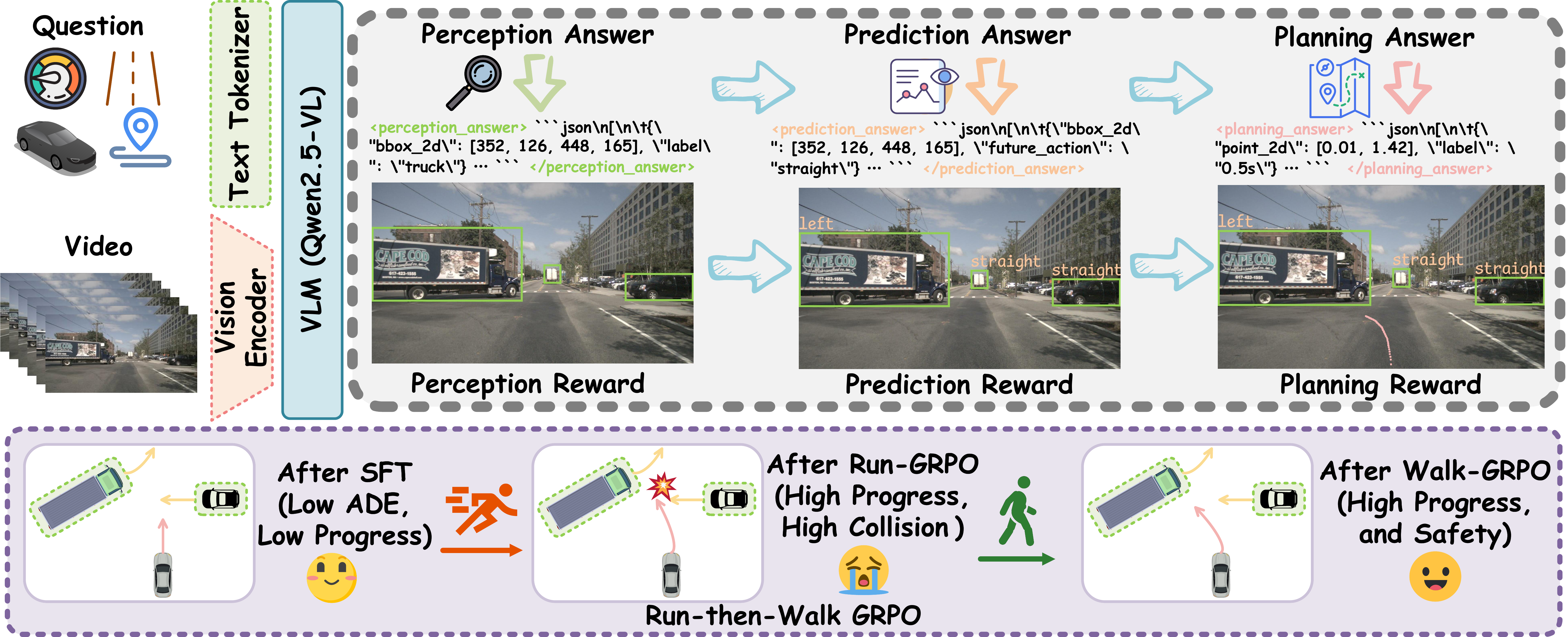}
\caption{\textbf{Illustrative Run-then-Walk scheduling strategy for AutoDrive-P$^3$ planner.} It performs RL in two stages: a \textit{Run} phase that discovers high-progress actions under relaxed constraints, followed by a \textit{Walk} phase that adds endpoint and safety rewards to produce a safe, high-progress policy.}
\label{fig:pipeline}
\vspace{-0.5cm}
\end{figure*}

\subsection{Run-then-Walk Strategy}
For each driving query $q$, we sample a group of $G$ complete planner outputs $\{x_i\}_{i=1}^{G}$ from the current policy and compute the stage-wise reward
\begin{equation}
R_i^{(s)}=
R_{\mathrm{aux}}(x_i)+
\lambda_{\mathrm{plan}} R_{\mathrm{plan}}^{(s)}(\tau_i),
\label{eq:total_reward}
\end{equation}
where $s\in\{\mathrm{run},\mathrm{walk}\}$ and $R_{\mathrm{aux}}$ collects any architecture-native rewards that remain fixed across stages. These can include format, perception, and prediction rewards for a structured autoregressive planner or the native auxiliary terms of a diffusion planner. 

The sampled rewards are normalized into group-relative advantages:
\begin{equation}
A_i^{(s)}=
\frac{R_i^{(s)}-\bar{R}^{(s)}}{\sigma_R^{(s)}+\epsilon_A},
\qquad
\bar{R}^{(s)}=\frac{1}{G}\sum_{j=1}^{G}R_j^{(s)},
\qquad
\sigma_R^{(s)}=\sqrt{\frac{1}{G}\sum_{j=1}^{G}\left(R_j^{(s)}-\bar{R}^{(s)}\right)^2}.
\end{equation}
We set $r_i=\pi_\theta(x_i|q)/\pi_{\theta_{\mathrm{old}}}(x_i|q)$, then optimize the policy with the GRPO objective:
\begin{equation}
\mathcal{J}^{(s)}(\theta)=
\mathbb{E}\!\left[
\frac{1}{G}\sum_{i=1}^{G}\min\!\big(r_iA_i^{(s)},\,
\mathrm{clip}(r_i,1-\epsilon,1+\epsilon)A_i^{(s)}\big)
\right]
-\beta_s\,\mathbb{E}\!\left[
D_{\mathrm{KL}}\!\left(\pi_\theta(\cdot|q)\,\|\,\pi_{\mathrm{ref}}^{(s)}(\cdot|q)\right)
\right].
\end{equation}

Starting from the SFT checkpoint $\pi_{\mathrm{sft}}$, we first run GRPO with $R_{\mathrm{plan}}^{\mathrm{run}}$ and use $\pi_{\mathrm{sft}}$ as the reference policy. This Run stage expands the progress distribution and discovers high-progress modes. We then select a Run checkpoint $\pi_{\mathrm{run}}$ and continue GRPO with $R_{\mathrm{plan}}^{\mathrm{walk}}$, using $\pi_{\mathrm{run}}$ as the new reference policy. The Walk stage therefore repairs safety around the progress-oriented behaviors discovered in Run, rather than collapsing back to the original conservative SFT solution.

\begin{table}[t]
\centering
\caption{\textbf{Performance comparison on NAVSIMv1 benchmark.}}
\vspace{-0.3cm}
\label{table:navsim}
\resizebox{1.0\linewidth}{!}{
\begin{tabular}{lcccccccc}
\toprule
\multicolumn{1}{c}{\textbf{Method}} & \textbf{Image} & \textbf{Lidar} & \textbf{NC}$\uparrow$ & \textbf{DAC}$\uparrow$ & \textbf{EP}$\uparrow$ & \textbf{TTC}$\uparrow$ & \textbf{Comf}$\uparrow$ & \cellcolor{lightbrown!50}\textbf{PDMS}$\uparrow$ \\
\midrule
Human         &\ding{55} & \ding{55}  & 100.0  & 100.0  & 87.5 & 100.0  & 99.9  & \cellcolor{lightbrown!50}94.8 \\
\midrule
Constant Velocity    &\ding{55} & \ding{55}    & 69.9 & 58.8 & 49.3 & 49.3 & 100.0 & \cellcolor{lightbrown!50}21.6 \\
Ego Status MLP      &\ding{55} & \ding{55}     & 93.0 & 77.3 & 62.8 & 83.6 & 100.0 & \cellcolor{lightbrown!50}65.6 \\
VADv2~\citep{weng2024drive}   & \ding{52} & \ding{55}  & 97.9 & 91.7 & 77.6 & 92.9 & 100.0 & \cellcolor{lightbrown!50}83.0 \\
UniAD~\citep{hu2023planning}  & \ding{52} & \ding{55}  & 97.8 & 91.9 & 78.8 & 92.9 & 100.0 & \cellcolor{lightbrown!50}83.4 \\
TransFuser~\citep{prakash2021multi} & \ding{52} & \ding{52} & 97.7 & 92.8 & 79.2 & 92.8 & 100.0 & \cellcolor{lightbrown!50}84.0 \\
PARA-Drive~\citep{weng2024drive} & \ding{52} & \ding{55} & 97.9 & 92.4 & 79.3 & 93.0 & 99.8  & \cellcolor{lightbrown!50}84.0 \\
Hydra-MDP~\citep{li2024hydra} & \ding{52} & \ding{52} & 98.3 & 96.0 & 78.7 & 94.6 & 100.0 & \cellcolor{lightbrown!50}86.5 \\
DiffusionDrive~\citep{liao2025diffusiondrive}  & \ding{52} & \ding{52}  & 98.2 & 96.2 & 82.2 & 94.7 & 100.0  & \cellcolor{lightbrown!50}88.1 \\
WoTE~\citep{li2025end}  & \ding{52} & \ding{52}  & 98.5 & 96.8 & 81.9 & 94.9 & 99.9  & \cellcolor{lightbrown!50}88.3 \\
DriveVLA-W0~\citep{li2025drivevla} & \ding{52} & \ding{55} & 98.7 & 99.1 & 83.3 & 95.3 & 99.3 & \cellcolor{lightbrown!50}90.2 \\
DriveFuture~\citep{hong2026drivefuture} & \ding{52} & \ding{55} & 99.1 & 97.2 & 84.5 & 96.0 & 100.0 & \cellcolor{lightbrown!50}90.7 \\
DriveWorld-VLA~\citep{liu2026driveworld} & \ding{52} & \ding{55} & 99.1 & 98.2 & 85.9 & 96.1 & 100.0 & \cellcolor{lightbrown!50}91.3 \\
\midrule
\midrule
ReCogDrive~\citep{li2025recogdrive} & \ding{52} & \ding{55} & 98.1 & 97.7 & 86.5 & 94.9 & 100.0 & \cellcolor{lightbrown!50}90.6 \\
ReCogDrive (Run-then-Walk, ours) & \ding{52} & \ding{55} & \textbf{98.6} & \textbf{97.9} & 85.9 & \textbf{95.8} & 100.0 & \cellcolor{lightbrown!50}\textbf{91.0} \\
\midrule
AutoDrive-P$^3$~\citep{ye2026autodrive} & \ding{52} & \ding{55} & 98.9 & 97.7 & 83.7 & 96.6 & 99.9 & \cellcolor{lightbrown!50}90.2 \\
AutoDrive-P$^3$ (Run-then-Walk, ours) & \ding{52} & \ding{55} & 98.6 & 97.3 & \textbf{88.9} & 95.5 & \textbf{100.0} & \cellcolor{lightbrown!50}\textbf{91.8} \\
\bottomrule
\end{tabular}}
\end{table}

\begin{table}[t]
\centering
\caption{\textbf{Performance comparison on NAVSIMv2 benchmark (Navtest).}}
\vspace{-0.3cm}
\label{table:navsimv2}
\resizebox{1.0\linewidth}{!}{
\begin{tabular}{lcccccccccc}
\toprule
\multicolumn{1}{c}{\textbf{Method}} & \textbf{NC}$\uparrow$ & \textbf{DAC}$\uparrow$ & \textbf{DDC}$\uparrow$ & \textbf{TLC}$\uparrow$ & \textbf{EP}$\uparrow$ & \textbf{TTC}$\uparrow$ & \textbf{LK}$\uparrow$ & \textbf{HC}$\uparrow$ & \textbf{EC}$\uparrow$ & \cellcolor{lightbrown!50}\textbf{EPDMS}$\uparrow$ \\
\midrule
Human & 100.0 & 100.0 & 99.8 & 100.0 & 87.4 & 100.0 & 100.0 & 98.1 & 90.1 & \cellcolor{lightbrown!50} 94.5 \\
\midrule
Ego Status MLP & 93.1 & 77.9 & 92.7 & 99.6 & 86.0 & 91.5 & 89.4 & 98.3 & 85.4 & \cellcolor{lightbrown!50} 64.0 \\
Transfuser~\citep{prakash2021multi} & 96.9 & 89.9 & 97.8 & 99.7 & 87.1 & 95.4 & 92.7 & 98.3 & 87.2 & \cellcolor{lightbrown!50} 84.0 \\
DiffusionDrive~\citep{liao2025diffusiondrive} & 98.2 & 96.2 & 99.5 & 99.8 & 87.4 & 97.3 & 96.9 & 98.4 & 87.7 & \cellcolor{lightbrown!50}88.2 \\
WoTE~\citep{li2025end} & 98.5 & 96.8 & 98.8 & 99.8 & 86.1 & 97.9 & 95.5 & 98.3 & 82.9 & \cellcolor{lightbrown!50}87.7 \\
DriveWorld-VLA~\citep{liu2026driveworld} & 98.6 & 99.1 & 99.6 & 99.8 & 87.4 & 97.9 & 97.0 & 97.8 & 78.6 & \cellcolor{lightbrown!50}86.8 \\
DriveVLA-W0~\citep{li2025drivevla} & 98.5 & 99.1 & 98.0 & 99.7 & 86.4 & 98.1 & 93.2 & 97.9 & 58.9 & \cellcolor{lightbrown!50} 86.1 \\
Latent-WAM~\citep{wang2026latent} & 98.1 & 97.3 & 99.6 & 99.8 & 87.7 & 97.3 & 97.6 & 98.1 & 87.3 & \cellcolor{lightbrown!50} 89.3 \\
\midrule
\midrule
ReCogDrive~\citep{li2025recogdrive} & 98.2 & 97.7 & 98.4 & 100.0 & 89.9 & 97.4 & 90.8 & 97.7 & 26.8 & \cellcolor{lightbrown!50}82.7 \\
ReCogDrive (Run-then-Walk, ours) & \textbf{98.5} & \textbf{97.9} & \textbf{98.8} & 100.0 & 89.4 & \textbf{97.8} & \textbf{92.0} & \textbf{98.2} & \textbf{29.8} & \cellcolor{lightbrown!50}\textbf{83.7} \\
\midrule
AutoDrive-P$^3$~\citep{ye2026autodrive} & 98.9 & 97.6 & 98.9 & 99.8 & 86.8 & 98.5 & 95.4 & 98.3 & 80.6 & \cellcolor{lightbrown!50} 88.7 \\
AutoDrive-P$^3$ (Run-then-Walk, ours) & 98.6 & 97.3 & 98.7 & 99.7 & \textbf{92.8} & 98.1 & \textbf{96.1} & 97.5 & 78.3 & \cellcolor{lightbrown!50}\textbf{89.6} \\
\bottomrule
\end{tabular}}
\vspace{-0.5cm}
\end{table}

\subsection{Idealized Theoretical Justification}
\label{sec:theory}

The following analysis is idealized and serves as a reward-direction explanation rather than a convergence guarantee. Run-then-Walk separates progress discovery from safety repair. On a fixed query distribution $q\sim\rho$, let $\pi(x\mid q)$ denote a complete planner-sample distribution and define
\begin{equation}
J(\pi)=\mathbb{E}_{\rho,\pi}[SQ],\qquad Q=\alpha P+(1-\alpha)B,\qquad 0\leq S,P,B\leq1,
\label{eq:theory_score}
\end{equation}
where $P$ is progress, $S$ contains multiplicative safety/compliance factors, and $B$ contains the remaining quality terms. This covers PDMS ($S=\mathrm{NC}\,\mathrm{DAC}$, $\alpha=5/12$) and EPDMS ($S=\mathrm{NC}\,\mathrm{DAC}\,\mathrm{DDC}\,\mathrm{TLC}$, $\alpha=5/16$). Since $J\leq\min\{\mathbb{E}[S],\mathbb{E}[Q]\}$, progress alone cannot compensate for poor safety.

In an idealized population update anchored at $\pi_R=\pi_{\mathrm{run}}$, maximizing $\mathbb{E}[U]-\beta D_{\mathrm{KL}}(\pi\|\pi_R)$ reweights samples by $e^{U/\beta}$. Walk sets $W=R_{\mathrm{plan}}^{\mathrm{walk}}$ to zero when the safety gate fails and gives $W\geq3/2$ when it passes. Thus a positive total-reward gap increases the probability of gate-passing samples. If the range of unchanged auxiliary rewards is at most $\omega$, a sufficient gap is $(3/2)\lambda_{\mathrm{plan}}-\omega>0$. This condition is stated per implementation and makes the argument independent of any particular planner architecture. 

If the achieved full-sample KL to $\pi_R$ is at most $\kappa$, Pinsker's inequality gives $\varepsilon_\kappa=\min\{1,\sqrt{\kappa/2}\}$ and, for $a_R=\Pr_R(P\geq p_0)$ and $f_W=\Pr_W(S<s_0)$,
\begin{equation}
\mathbb{E}_W[P]\geq\mathbb{E}_R[P]-\varepsilon_\kappa,\qquad
J(\pi_W)\geq\alpha p_0s_0[a_R-\varepsilon_\kappa-f_W]_+.
\label{eq:theory_lower}
\end{equation}
Run must therefore create high-progress mass, Walk must reduce safety failures, and the reference must retain enough of that mass. Writing $s_t=\mathbb{E}_t[S]$ and $c_t=\mathbb{E}_t[SQ]/s_t$, a quality-loss bound $c_W\geq c_R-\ell$ yields
\begin{equation}
J(\pi_W)-J(\pi_R)\geq(s_W-s_R)c_R-s_W\ell,
\label{eq:theory_gain}
\end{equation}
so safety improvement must exceed quality lost among safety-weighted trajectories.

These statements require positive Run support for useful safe-progress samples, adequate group coverage, a fixed evaluation distribution, and a measured rather than nominal KL bound. A zero KL forbids repair, and our experiments further show that removing the KL term leads to training collapse; excessive KL weakens progress retention. The Walk gate is not identical to EPDMS safety, and endpoint proximity is neither a path-safety nor a progress certificate. Full derivations, the idealized probability-ratio bound, and finite-sampling conditions are provided in Appendix~\ref{app:theory}.

\begin{table*}[t]
\centering
\caption{\textbf{Performance comparison on Navhard benchmark.}}
\vspace{-0.3cm}
\label{table:navhard}
\resizebox{\textwidth}{!}{%
\begin{tabular}{l>{\columncolor{lightbrown!50}}c cccccccccc}
\toprule
\textbf{Methods} & \textbf{EPDMS$\uparrow$} & \textbf{Stage} & \textbf{NC$\uparrow$} & \textbf{DAC$\uparrow$} & \textbf{DDC$\uparrow$} & \textbf{TLC$\uparrow$} & \textbf{EP$\uparrow$} & \textbf{TTC$\uparrow$} & \textbf{LK$\uparrow$} & \textbf{HC$\uparrow$} & \textbf{EC$\uparrow$} \\
\midrule
\multirow{2}{*}{LTF~\citep{chitta2022transfuser}} & & 1 & 96.2 & 79.5 & 99.1 & 99.5 & 84.1 & 95.1 & 94.2 & 97.5 & 79.1 \\
                         & \multirow[c]{-2}{*}{23.1} & 2 & 77.7 & 70.2 & 84.2 & 98.0 & 85.1 & 75.6 & 45.4 & 95.7 & 75.9 \\
\multirow{2}{*}{DiffusionDrive~\citep{liao2025diffusiondrive}} & & 1 & 96.8 & 88.2 & 99.3 & 99.3 & 84.5 & 94.6 & 95.5 & 97.5 & 79.1 \\
                                   & \multirow[c]{-2}{*}{28.9} & 2 & 80.3 & 74.4 & 86.1 & 98.4 & 87.4 & 76.9 & 50.4 & 95.5 & 69.4 \\
\multirow[c]{2}{*}{GoalFlow~\citep{Xing_2025_CVPR}} & & 1 & 96.0 & 92.6 & 99.3 & 99.3 & 84.0 & 95.7 & 97.1 & 97.5 & 40.4 \\
                              & \multirow[c]{-2}{*}{28.7} & 2 & 79.4 & 78.2 & 86.4 & 97.7 & 86.5 & 76.0 & 45.5 & 94.4 & 40.4 \\
\midrule
\midrule
\multirow{2}{*}{ReCogDrive \citep{li2025recogdrive}} & & 1 & 95.4 & 90.0 & 96.0 & 99.6 & 86.3 & 94.4 & 90.9 & 97.1 & 28.9 \\
                          & \multirow[c]{-2}{*}{26.5} & 2 & 76.8 & 72.6 & 80.4 & 98.3 & 90.5 & 73.8 & 42.2 & 95.6 & 34.7 \\
\multirow{2}{*}{ReCogDrive (Run-then-Walk, ours)} & & 1 & \textbf{95.6} & \textbf{91.3} & \textbf{96.8} & \textbf{99.8} & 85.9 & \textbf{95.3} & \textbf{92.2} & \textbf{97.8} & \textbf{32.9} \\
                     & \multirow[c]{-2}{*}{\textbf{28.2}} & 2 & \textbf{78.3} & \textbf{73.3} & \textbf{81.0} & \textbf{98.8} & 89.2 & \textbf{75.0} & \textbf{43.0} & \textbf{95.8} & \textbf{36.7} \\
\midrule
\multirow{2}{*}{AutoDrive-P$^3$ \citep{ye2026autodrive}} & & 1 & 96.8 & 88.7 & 97.1 & 99.6 & 84.0 & 95.1 & 93.1 & 97.6 & 68.4 \\
                              & \multirow[c]{-2}{*}{29.3} & 2 & 81.2 & 68.3 & 82.7 & 98.4 & 86.4 & 77.8 & 42.3 & 95.4 & 62.4 \\
\multirow{2}{*}{AutoDrive-P$^3$ (Run-then-Walk, ours)} & & 1 & \textbf{97.1} & 86.9 & 96.9 & 99.3 & \textbf{89.9} & \textbf{95.8} & \textbf{96.9} & 97.6 & 61.8 \\
                                  & \multirow[c]{-2}{*}{\textbf{31.3}} & 2 & \textbf{81.6} & \textbf{72.1} & 81.5 & 98.0 & \textbf{91.5} & 76.3 & \textbf{50.1} & 95.0 & 52.0 \\
\bottomrule
\end{tabular}}
\end{table*}

\begin{table}[t]
\centering
\caption{\textbf{Performance comparison on nuScenes Benchmark.}}
\vspace{-0.3cm}
\label{table:nuscenes}
\resizebox{1.0\linewidth}{!}{%
\begin{tabular}{lcccccccccc}
\toprule
\multicolumn{1}{c}{\multirow{2}{*}[-1ex]{\textbf{Method}}} & \multicolumn{4}{c}{\textbf{L2 (m)} $\downarrow$} & \multicolumn{4}{c}{\textbf{Collision} (\%) $\downarrow$} & \multirow{2}{*}[-1ex]{\textbf{VLM}} \\
\cmidrule(lr){2-5}\cmidrule(lr){6-9}
 & 1s & 2s & 3s & \cellcolor{lightbrown!50}Avg. & 1s & 2s & 3s & \cellcolor{lightbrown!50}Avg. & \\
\midrule
ST-P3~\citep{hu2022st} & 1.33 & 2.11 & 2.90 & \cellcolor{lightbrown!50}2.11 & 0.23 & 0.62 & 1.27 & \cellcolor{lightbrown!50}0.71 & - \\
VAD~\citep{jiang2023vad} & 0.17 & 0.34 & 0.60 & \cellcolor{lightbrown!50}0.37 & 0.07 & 0.10 & 0.24 & \cellcolor{lightbrown!50}0.14 & - \\
Ego-MLP~\citep{li2024ego} & 0.46 & 0.76 & 1.12 & \cellcolor{lightbrown!50}0.78 & 0.21 & 0.35 & 0.58 & \cellcolor{lightbrown!50}0.38 & - \\
UniAD~\citep{hu2023planning} & 0.44 & 0.67 & 0.96 & \cellcolor{lightbrown!50}0.69 & 0.04 & 0.08 & 0.23 & \cellcolor{lightbrown!50}0.12 & - \\
InsightDrive~\citep{song2025insightdrive} & 0.23 & 0.41 & 0.68 & \cellcolor{lightbrown!50}0.44 & 0.09 & 0.10 & 0.27 & \cellcolor{lightbrown!50}0.15 & - \\
\midrule
GPT-Driver~\citep{mao2023gpt} & 0.20 & 0.40 & 0.70 & \cellcolor{lightbrown!50}0.44 & 0.04 & 0.12 & 0.36 & \cellcolor{lightbrown!50}0.17 & GPT-3.5 \\
DriveVLM~\citep{tian2024drivevlm} & 0.18 & 0.34 & 0.68 & \cellcolor{lightbrown!50}0.40 & 0.10 & 0.22 & 0.45 & \cellcolor{lightbrown!50}0.27 & Qwen2-VL-7B \\
OpenEMMA~\citep{xing2025openemma} & 1.45 & 3.21 & 3.76 & \cellcolor{lightbrown!50}2.81 & -- & -- & -- & \cellcolor{lightbrown!50}-- & Qwen2-VL-7B \\
RDA-Driver~\citep{huang2024making} & 0.17 & 0.37 & 0.69 & \cellcolor{lightbrown!50}0.40 & 0.01 & 0.05 & 0.26 & \cellcolor{lightbrown!50}0.10 & LLaVA-7B \\
OmniDrive~\citep{wang2024omnidrive} & 0.14 & 0.29 & 0.55 & \cellcolor{lightbrown!50}0.33 & 0.01 & 0.04 & 0.27 & \cellcolor{lightbrown!50}0.11 & LLaVA-7B \\
OpenDriveVLA~\citep{zhou2025opendrivevla} & 0.14 & 0.30 & 0.55 & \cellcolor{lightbrown!50}0.33 & 0.02 & 0.07 & 0.22 & \cellcolor{lightbrown!50}0.10 & Qwen2.5-VL-3B \\
AutoVLA~\citep{zhou2025autovla} & 0.25 & 0.46 & 0.73 & \cellcolor{lightbrown!50}0.48 & 0.07 & 0.07 & 0.26 & \cellcolor{lightbrown!50}0.13 & Qwen2.5-VL-3B \\
AutoDrive-R$^2$~\citep{yuan2025autodrive} & 0.35 & 0.49 & 0.62 & \cellcolor{lightbrown!50}0.49 & -- & -- & -- & \cellcolor{lightbrown!50}-- & Qwen2.5-VL-3B \\
\midrule
\midrule
AutoDrive-P$^3$~\citep{ye2026autodrive} & 0.16 & 0.31 & 0.56 & \cellcolor{lightbrown!50}0.34 & \textbf{0.00} & 0.04 & 0.20 & \cellcolor{lightbrown!50}0.08 & Qwen2.5-VL-3B \\
AutoDrive-P$^3$ (Run-then-Walk, ours) & \textbf{0.15} & \textbf{0.30} & \textbf{0.55} & \cellcolor{lightbrown!50}\textbf{0.33} & 0.01 & \textbf{0.03} & \textbf{0.16} & \cellcolor{lightbrown!50}\textbf{0.06} & Qwen2.5-VL-3B \\
\bottomrule
\end{tabular}}
\end{table}

\vspace{-0.25cm}
\section{Experiments}
\label{sec:experiments}
\vspace{-0.25cm}
\textbf{Benchmarks and baselines.}
We evaluate Run-then-Walk on four benchmarks: NAVSIMv1~\citep{dauner2024navsim}, NAVSIMv2~\citep{cao2025pseudo}, Navhard~\citep{cao2025pseudo}, and nuScenes~\citep{caesar2020nuscenes}. To validate the robustness of our method, we evaluate it on two representative VLM planners: the autoregressive AutoDrive-P$^3$ (fast-mode) and the diffusion-based ReCogDrive. NAVSIMv1 uses the Predictive Driver Model Score (PDMS), whereas NAVSIMv2 and Navhard use the Extended PDMS (EPDMS). nuScenes provides trajectory-error and collision evaluation. Since only AutoDrive-P$^3$ has been trained on nuScenes, we restrict the nuScenes comparison to this planner. We set $R_{\mathrm{agg}}$ to PDMS, and for the endpoint reward, we set $\Delta=10$ and $\eta=0.2$ in all experiments.

\textbf{Main results across planners and benchmarks.}
Tables~\ref{table:navsim} and~\ref{table:navsimv2} show consistent improvements for both planner families rather than a gain tied to one base model. On NAVSIMv1, Run-then-Walk raises the paired diffusion-planner PDMS from $90.6$ to $91.0$ and the paired autoregressive-planner PDMS from $90.2$ to $91.8$. The ReCogDrive comparison improves all safety-related metrics, only with a small decrease in EP, whereas the AutoDrive-P$^3$ comparison highlights increased progress while retaining strong safety. For AutoDrive-P$^3$, the schedule uses 3 Run epochs and 2 Walk epochs, for 5 RL epochs in total. For ReCogDrive, we use 5 Run epochs and 1 Walk epoch, for 6 RL epochs in total. Thus the method saves 40--50\% of the 10-epoch baseline training steps, converges faster, and improves performance for both planners. On NAVSIMv2 (Table~\ref{table:navsimv2}), paired EPDMS rises from $82.7$ to $83.7$ for ReCogDrive and from $88.7$ to $89.6$ for AutoDrive-P$^3$ with EP $92.8$. On the two-stage Navhard protocol, as shown in Table~\ref{table:navhard}, paired aggregates improve from $26.5$ to $28.2$ and from $29.3$ to $31.3$, respectively. On nuScenes, as shown in Table~\ref{table:nuscenes}, AutoDrive-P$^3$ planner reduces average trajectory error and collision rate. Across different benchmarks and planner methods, our Run-then-Walk strategy not only better balances safety and progress, thereby improving driving performance, but also substantially saves training time.

\vspace{-0.25cm}
\section{Analysis and Visualization}
\label{sec:analysis}
\vspace{-0.25cm}

\begin{figure*}[t]
\centering
\includegraphics[width=0.9\textwidth]{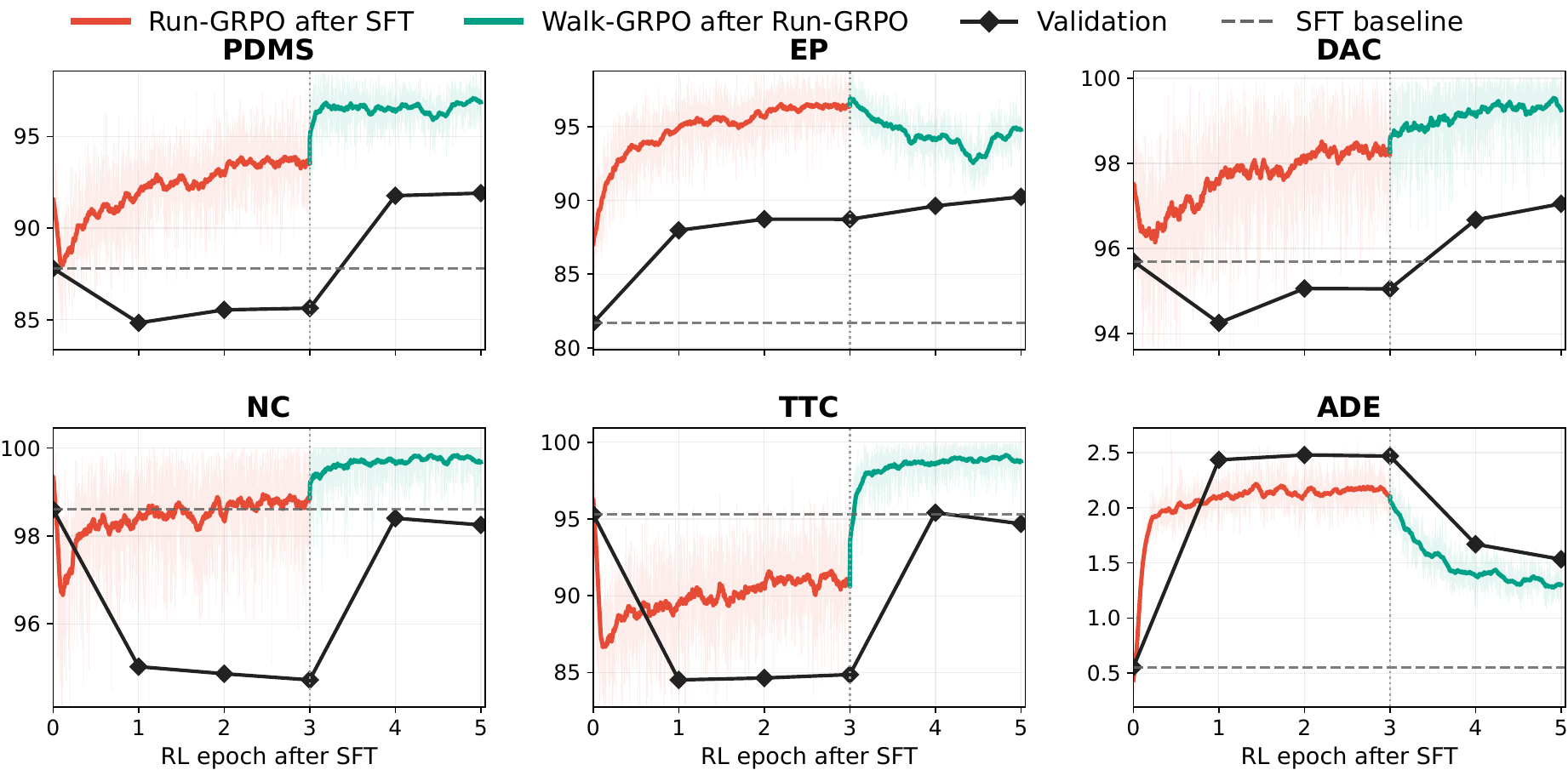}
\vspace{-0.2cm}
\caption{\textbf{PDMS sub-metric evolution during AutoDrive-P$^3$ Run-then-Walk strategy.} Red curves denote the 3-epoch Run-GRPO after SFT, green curves denote the following 2-epoch Walk-GRPO after Run-GRPO. Run-GRPO increases progress but hurts safety; Walk-GRPO then restores safety, lifts PDMS above the SFT baseline, and yields a safe, high-progress policy in only 5 epochs.}
\label{fig:run_then_walk_metrics}
\vspace{-0.4cm}
\end{figure*}

\begin{table}[t]
\centering
\begin{minipage}[t]{0.49\linewidth}
\centering
\footnotesize\caption{\textbf{Run/Walk recipe comparison.}}
\vspace{-0.3cm}
\label{table:rl_recipes}
\resizebox{\linewidth}{!}{
\begin{tabular}{lcccccc}
\toprule
\multicolumn{1}{c}{\textbf{Different RL recipes}} & \textbf{NC}$\uparrow$ & \textbf{DAC}$\uparrow$ & \textbf{EP}$\uparrow$ & \textbf{TTC}$\uparrow$ & \textbf{Comf}$\uparrow$ & \cellcolor{lightbrown!50}\textbf{PDMS}$\uparrow$ \\
\midrule
Only Run  & 94.8 & 95.1 & \underline{88.8} & 84.8 & \textbf{100.0} & \cellcolor{lightbrown!50}85.7 \\
Only Walk  & \textbf{99.1} & \textbf{97.7} & 81.5 & \textbf{97.3} & \textbf{100.0} & \cellcolor{lightbrown!50}89.7 \\
Walk-and-Run  & 98.0 & 94.2 & 85.4 & 94.0 & \textbf{100.0} & \cellcolor{lightbrown!50}85.9 \\
Walk-then-Run  & \underline{98.9} & \textbf{97.7} & 83.7 & 96.6 & \underline{99.9} & \cellcolor{lightbrown!50}\underline{90.2} \\
\midrule
\textbf{Ours (Run-then-Walk)}  & \textbf{98.6} & \textbf{97.3} & \textbf{88.9} & \underline{95.5} & \textbf{100.0} & \cellcolor{lightbrown!50}\textbf{91.8} \\ 
\bottomrule
\end{tabular}
}
\end{minipage}
\hfill
\begin{minipage}[t]{0.49\linewidth}
\centering
\footnotesize\caption{\textbf{Effect of safety metrics in \textit{Walk}.}}
\vspace{-0.3cm}
\label{table:safety_metrics}
\resizebox{\linewidth}{!}{
\begin{tabular}{ccc|cccccc}
\toprule
\multicolumn{3}{c|}{\textbf{Reward Signals}} & \multicolumn{6}{c}{\textbf{Evaluation Metrics}} \\
\cmidrule(lr){1-3} \cmidrule(lr){4-9}
\textbf{DAC} & \textbf{NC} & \textbf{TTC} & \textbf{NC}$\uparrow$ & \textbf{DAC}$\uparrow$ & \textbf{EP}$\uparrow$ & \textbf{TTC}$\uparrow$ & \textbf{Comf}$\uparrow$ & \cellcolor{lightbrown!50}\textbf{PDMS}$\uparrow$ \\
\midrule
\ding{52} & \ding{55} & \ding{55} & 98.5 & 97.1 & \underline{88.0} & 94.9 & 100.0 & \cellcolor{lightbrown!50}\underline{91.5} \\
\ding{52} & \ding{52} & \ding{55} & \underline{98.6} & \underline{97.3} & \textbf{88.9} & \underline{95.5} & 100.0 & \cellcolor{lightbrown!50}\textbf{91.8} \\
\ding{52} & \ding{52} & \ding{52} & \textbf{98.7} & \textbf{97.4} & 86.9 & \textbf{96.0} & 100.0 & \cellcolor{lightbrown!50}91.2 \\
\bottomrule
\end{tabular}
}
\end{minipage}
\vspace{-0.5cm}
\end{table}

\textbf{Analysis of Run-then-Walk Training Strategy.}
Figure~\ref{fig:run_then_walk_metrics} reveals the behavioral evolution of AutoDrive-P$^3$ across the two training stages.
During \textit{Run}-GRPO, the EP rises substantially (81.7$\to$88.8), but the safety metrics drop, indicating that the policy learns to maximize progress at the cost of frequent collisions and unsafe events, ultimately degrading PDMS from 87.8 to 85.7.
In the subsequent \textit{Walk}-GRPO, safety is strongly restored, while the high EP is preserved, and the final checkpoint reaches 91.8 PDMS.
This AutoDrive-P$^3$ result uses only 5 RL epochs, and the same safety-repair principle applied to ReCogDrive reaches 91.0 PDMS in 6 epochs, together saving 40--50\% of RL training steps relative to the 10-epoch baselines while improving performance. More detailed analysis, including ReCogDrive, is provided in the supplementary material.

\textbf{Analysis of different RL recipes.}
Table~\ref{table:rl_recipes} compares different orderings of Run and Walk objectives.
\textit{Only Run} yields high EP ($88.8$) but the lowest PDMS ($85.7$) due to poor safety; \textit{Only Walk} achieves the best safety but limits progress (EP $81.5$, PDMS $89.7$).
\textit{Walk-then-Run} is suboptimal, whereas our \textit{Run-then-Walk} achieves the highest PDMS by exploring progress first and restoring safety afterward.

\textbf{Effect of safety signal composition.}
Table~\ref{table:safety_metrics} ablates which safety signals are used in the Walk-phase planning reward $R_{\mathrm{plan}}^{\mathrm{walk}}$ (Eq.~\ref{eq:walk_plan_reward}). Using only DAC yields 91.5 PDMS. Adding NC improves the balance to 91.8 PDMS and 88.9 EP. Further adding TTC raises safety scores but lowers EP to 86.9, reducing PDMS to 91.2. These results suggest that DAC+NC provides the best safety--progress trade-off, while TTC as a reward signal introduces over-conservatism.

\begin{figure*}[t]
\centering
\begin{subfigure}[t]{0.9\linewidth}
\centering
\includegraphics[width=0.9\linewidth]{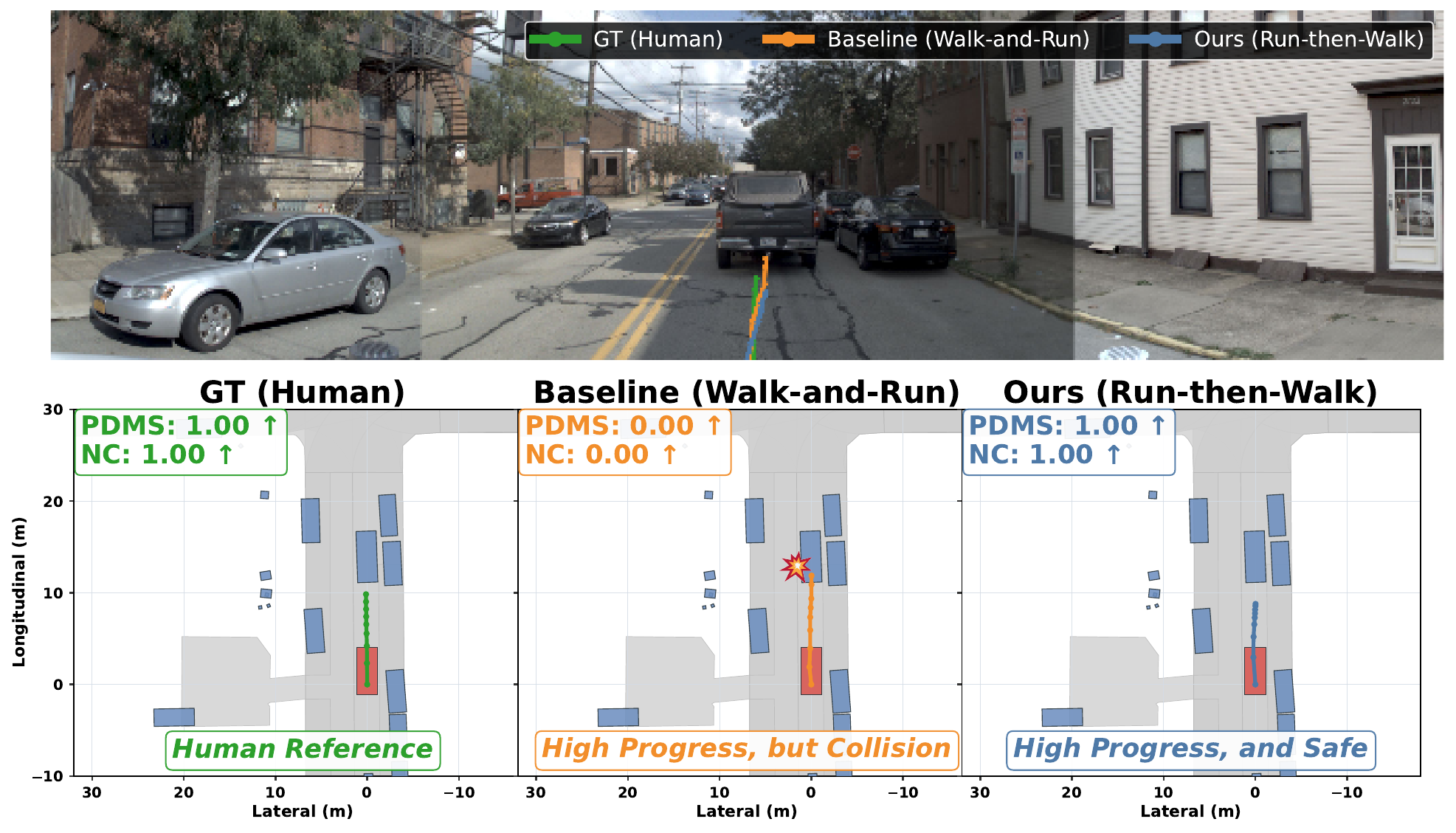}
\label{fig:vis_recogdrive}
\end{subfigure}
\vspace{-0.3cm}
\begin{subfigure}[t]{0.9\linewidth}
\centering
\includegraphics[width=0.9\linewidth]{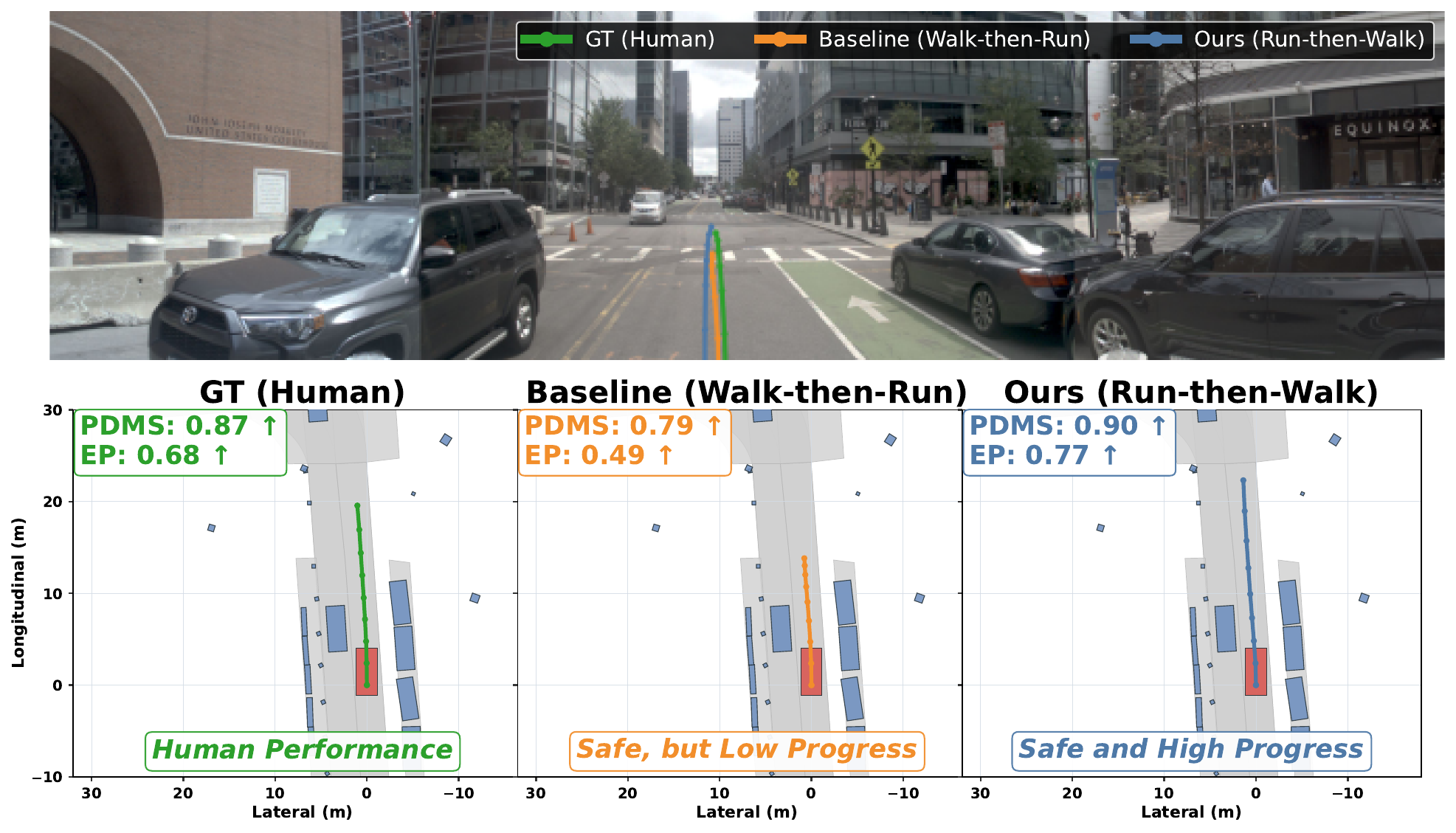}
\label{fig:vis_autodrive}
\end{subfigure}
\caption{\textbf{Qualitative trajectory visualization.}}
\label{fig:qualitative_vis}
\vspace{-0.25cm}
\end{figure*}

\textbf{Visualization.} Figure~\ref{fig:qualitative_vis} presents qualitative trajectory comparisons on two representative driving scenarios. Walk-and-Run achieves high progress but collides (NC $=0$), and Walk-then-Run remains safe but advances less (EP $=0.65$). Our method maintains a high EP while ensuring safety.

\vspace{-0.15cm}
\section{Conclusion}
\label{sec:conclusion}
\vspace{-0.15cm}

In this paper, we first reveal two RL training recipe regimes: a progress regime (Run-GRPO) that aggressively explores high progress and a safety regime (Walk-GRPO) that restores safety. Then, we present \textit{Run-then-Walk}, a simple and effective two-stage RL training scheduling strategy for VLM-based autonomous driving. By first exploring progress under relaxed constraints and then repairing safety with endpoint rewards, our method escapes SFT conservatism while avoiding the unsafe driving. Across VLM-based autoregressive and diffusion planners on different benchmarks, experiments show improved driving performance and a 40--50\% reduction in RL training steps. These findings demonstrate that a progress-first, safety-refinement curriculum is efficient and effective, and we hope this simple paradigm inspires future RL in broader embodied decision-making.

\bibliography{iclr2027_conference}
\bibliographystyle{iclr2027_conference}

\clearpage
\appendix

\section{Reward Function Details}
\label{app:reward_details}

\subsection{Planner-Agnostic Reward Interface}

For each driving query $q$, we sample a group of $G$ complete planner outputs
$\{x_i\}_{i=1}^{G}$ from $\pi_\theta$ and decode each output into a trajectory
$\tau_i$. An autoregressive planner assigns the resulting reward to the sampled
token sequence, whereas a diffusion planner assigns it to the sampled continuous
trajectory. In both cases, the total reward is
\begin{equation}
R_i^{(s)}=R_{\mathrm{aux}}(x_i)+\lambda_{\mathrm{plan}}
R_{\mathrm{plan}}^{(s)}(\tau_i),\qquad
s\in\{\mathrm{run},\mathrm{walk}\}.
\label{eqa:mix_reward}
\end{equation}
Here $R_{\mathrm{aux}}$ denotes any native auxiliary objectives of the underlying
planner. These objectives and the output representation remain unchanged between
the two stages. Rewards are normalized within each sampled group and optimized
with the GRPO objective in Section~\ref{sec:method}. Consequently, the only
algorithmic change shared by both planner families is the planning reward and its
Run-to-Walk schedule.

\subsection{Planning Reward: Run vs.\ Walk}
\label{app:planning_reward}

The planning reward is the only reward component that differs between the Run and Walk stages.

\textbf{Run Phase Planning Reward.}
In the Run phase, we directly use the closed-loop PDMS as the planning reward:
\begin{equation}
R_{\mathrm{plan}}^{\mathrm{run}}(\tau_i) = \mathrm{PDMS}(\tau_i).
\label{eqa:run_reward}
\end{equation}
No endpoint/L2 constraint or explicit safety reward is applied. This intentionally encourages the policy to maximize progress under relaxed constraints.

\textbf{Walk Phase Planning Reward.}
In the Walk phase, we replace the PDMS-based planning reward with a composite of endpoint and safety rewards. The endpoint reward is defined via the $L_1$ distance $d_E = \|p_T - g_T\|_1$ between the final predicted point $p_T$ and the expert endpoint $g_T$:
\begin{equation}
k(d_E)=\min\!\left(K,\;\max\!\left(0,\left\lfloor (d_E-2\Delta)/\Delta\right\rfloor + 1\right)\right), \qquad R_{\mathrm{end}}(d_E)=1-\eta\,k(d_E),
\end{equation}
where $\Delta>0$ is the step size, $\eta \in (0,1)$ is the reward decrement, and $K=\lfloor 1/\eta \rfloor$. We use the fixed configuration $\Delta=10$, $\eta=0.2$ throughout. The complete Walk planning reward is:
\begin{equation}
R_{\mathrm{plan}}^{\mathrm{walk}}(\tau_i,\tau_i^\star) = \mathbb{I}_{\mathrm{safe}}(\tau_i)\,\Big(\mathrm{NC}(\tau_i)+\mathrm{DAC}(\tau_i)+R_{\mathrm{end}}(d_E)\Big),
\label{eqa:walk_reward}
\end{equation}
where $\mathbb{I}_{\mathrm{safe}}(\tau_i)=1$ only if $\mathrm{NC}(\tau_i)>0$, and $\mathrm{DAC}(\tau_i)>0$; otherwise the entire planning reward is zeroed. Algorithm~\ref{alg:rtw_full} presents the complete Run-then-Walk scheduling strategy procedure.



\begin{algorithm}[t]
\caption{Run-then-Walk training scheduling strategy}
\label{alg:rtw_full}
\begin{algorithmic}[1]
\Require SFT policy $\pi_{\mathrm{sft}}$, queries $\mathcal{Q}$, group size $G$, auxiliary reward $R_{\mathrm{aux}}$, planning weight $\lambda_{\mathrm{plan}}$, KL coefficients $\beta_{\mathrm{run}},\beta_{\mathrm{walk}}$
\Ensure Optimized driving policy $\pi_{\mathrm{walk}}$
\Statex
\Statex \textbf{--- Stage 1: Run-GRPO (progress discovery) ---}
\State Initialize $\pi \leftarrow \pi_{\mathrm{sft}}$, set reference $\pi_{\mathrm{ref}} \leftarrow \pi_{\mathrm{sft}}$
\For{each query $q \in \mathcal{Q}$}
    \State Sample outputs $\{x_i\}_{i=1}^{G} \sim \pi(\cdot\mid q)$ and decode trajectories $\{\tau_i\}_{i=1}^{G}$
    \For{$i=1$ to $G$}
        \State {$R_{\mathrm{plan}}^i \leftarrow R_{\mathrm{plan}}^{run}(\tau_i)$} \Comment{Focus on progress}
        \State $R_i \leftarrow R_{\mathrm{aux}}(x_i)+\lambda_{\mathrm{plan}}R_{\mathrm{plan}}^i$
    \EndFor
    \State Compute advantages $\hat{A}_i = (R_i - \bar{R})/\sigma_R$ and update $\pi$ via GRPO with KL penalty $\beta_{\mathrm{run}}$
\EndFor
\State Select Run checkpoint $\pi_{\mathrm{run}}$
\Statex
\Statex \textbf{--- Stage 2: Walk-GRPO (safety repair) ---}
\State Initialize $\pi \leftarrow \pi_{\mathrm{run}}$, set reference $\pi_{\mathrm{ref}} \leftarrow \pi_{\mathrm{run}}$
\For{each query $q \in \mathcal{Q}$}
    \State Sample outputs $\{x_i\}_{i=1}^{G} \sim \pi(\cdot\mid q)$ and decode trajectories $\{\tau_i\}_{i=1}^{G}$
    \For{$i=1$ to $G$}
        \State {$R_{\mathrm{plan}}^i \leftarrow R_{\mathrm{plan}}^{\mathrm{walk}}(\tau_i,\tau_i^\star)$} \Comment{Endpoint + safety gate}
        \State $R_i \leftarrow R_{\mathrm{aux}}(x_i)+\lambda_{\mathrm{plan}}R_{\mathrm{plan}}^i$
    \EndFor
    \State Compute advantages $\hat{A}_i = (R_i - \bar{R})/\sigma_R$ and update $\pi$ via GRPO with KL penalty $\beta_{\mathrm{walk}}$
\EndFor
\State \Return $\pi_{\mathrm{walk}}$
\end{algorithmic}
\end{algorithm}

\section{Idealized Theoretical Justification}
\label{app:theory}

This appendix provides the full derivations for the idealized theoretical justification summarized in Section~\ref{sec:theory}. They are conditional statements about a fixed query distribution, not a claim that every finite-sample GRPO run must improve. Let $q\sim\rho$ and let $x$ be a complete planner sample (an autoregressive response or a diffusion sample) with induced trajectory $\tau(x)$. We write
\begin{equation}
J(\pi)=\mathbb{E}_{\rho,\pi}[S Q],\qquad Q=\alpha P+(1-\alpha)B,\qquad 0\leq S,P,B\leq1,
\label{eq:app_theory_score}
\end{equation}
where $P$ is normalized progress, $S$ is the product of safety/compliance factors, and $B$ contains the remaining quality terms. For PDMS, $S=\mathrm{NC}\,\mathrm{DAC}$ and $\alpha=5/12$; for EPDMS, $S=\mathrm{NC}\,\mathrm{DAC}\,\mathrm{DDC}\,\mathrm{TLC}$ and $\alpha=5/16$. Since $0\leq S,Q\leq1$, $J\leq\mathbb{E}[S]$ and $J\leq\mathbb{E}[Q]$. This is the basic progress--safety bottleneck: high progress cannot compensate for a vanishing safety factor.

\paragraph{Safety direction of the Walk reward.}
Fix $q$ and let $\pi_R=\pi_{\mathrm{run}}$, $H=\mathbb{I}_{\mathrm{safe}}$, and $W=R_{\mathrm{plan}}^{\mathrm{walk}}$. Let $C$ denote the unchanged architecture-native auxiliary reward and $U=C+\lambda_{\mathrm{plan}}W$. Consider the idealized population problem
\begin{equation}
\pi^\dagger=\arg\max_\pi\left\{\mathbb{E}_\pi[U]-\beta D_{\mathrm{KL}}(\pi\|\pi_R)\right\},\qquad
\pi^\dagger(x\mid q)=\frac{\pi_R(x\mid q)e^{U(q,x)/\beta}}{Z_q},
\label{eq:app_theory_tilt}
\end{equation}
where $Z_q=\mathbb{E}_{\pi_R}[e^{U/\beta}]$ and $\beta>0$. The variational identity
\begin{equation}
\mathbb{E}_\pi[U]-\beta D_{\mathrm{KL}}(\pi\|\pi_R)
=\beta\log Z_q-\beta D_{\mathrm{KL}}(\pi\|\pi^\dagger)
\end{equation}
proves the optimizer. Suppose both values of $H$ have positive reference probability and every passing sample has total reward at least $\gamma>0$ larger than every failing sample. If $h_R=\Pr_{\pi_R}(H=1\mid q)$ and $h^\dagger=\Pr_{\pi^\dagger}(H=1\mid q)$, then
\begin{equation}
\frac{h^\dagger}{1-h^\dagger}
=\frac{h_R}{1-h_R}
\frac{\mathbb{E}_{\pi_R}[e^{U/\beta}\mid H=1]}{\mathbb{E}_{\pi_R}[e^{U/\beta}\mid H=0]}
\geq e^{\gamma/\beta}\frac{h_R}{1-h_R},
\label{eq:app_theory_safety}
\end{equation}
and therefore
\begin{equation}
1-h^\dagger\leq\frac{1-h_R}{1-h_R+h_Re^{\gamma/\beta}}.
\end{equation}

The Walk construction supplies a concrete sufficient reward gap. Recall that
\[
R_{\mathrm{plan}}^{\mathrm{walk}}(\tau_i,\tau_i^\star)
=
\mathbb{I}_{\mathrm{safe}}(\tau_i)\,
\Big(\mathrm{NC}(\tau_i)+\mathrm{DAC}(\tau_i)+R_{\mathrm{end}}(d_E)\Big),
\]
where $\mathrm{NC}\in\{0,1/2,1\}$, $\mathrm{DAC}\in\{0,1\}$, and $R_{\mathrm{end}}\in[0,1]$. When the safety gate passes, we have $\mathrm{NC}>0$ and $\mathrm{DAC}>0$; hence $\mathrm{NC}\geq1/2$, $\mathrm{DAC}=1$, and $R_{\mathrm{end}}\geq0$. Therefore
\[
W=R_{\mathrm{plan}}^{\mathrm{walk}}\geq \frac{1}{2}+1+0=\frac{3}{2}.
\]
When the gate fails, $\mathbb{I}_{\mathrm{safe}}=0$ and thus $W=0$. Let $C$ denote the unchanged auxiliary reward, whose range at a fixed query is at most $\omega$, and let the total reward be $U=C+\lambda_{\mathrm{plan}}W$. Then every passing sample satisfies
\[
U_{\text{pass}}\geq C_{\min}+\frac{3}{2}\lambda_{\mathrm{plan}},
\]
while every failing sample satisfies
\[
U_{\text{fail}}\leq C_{\max}.
\]
Hence the reward gap is at least
\[
U_{\text{pass}}-U_{\text{fail}}
\geq \frac{3}{2}\lambda_{\mathrm{plan}}-(C_{\max}-C_{\min})
\geq \frac{3}{2}\lambda_{\mathrm{plan}}-\omega.
\]
Thus $\gamma=(3/2)\lambda_{\mathrm{plan}}-\omega>0$ is sufficient for all passing samples to dominate all failing samples in total reward. The value of $\omega$ must be established from the actual normalization and scale of the auxiliary rewards for the implementation under study; it is not fixed by the gate construction alone.

The idealized result describes the local direction of group-normalized GRPO as well. At the Run checkpoint, clipping is inactive and the KL term has zero first derivative. Along $\pi_\zeta(x\mid q)\propto\pi_R(x\mid q)e^{\zeta H(x)}$, the derivative of the sampled surrogate at $\zeta=0$ is $G^{-1}\sum_i A_iH_i$, where $A_i$ are fixed Walk advantages and $H_i=H(x_i)$. Since $\sum_iA_i=0$ and all passing rewards exceed all failing rewards, this derivative is nonnegative and is positive when a group contains both outcomes. Shared neural parameters, finite groups, advantage recomputation, and clipping mean that the actual update need not realize this direction or the Gibbs optimizer. The parameter $\beta$ in Eq.~(\ref{eq:app_theory_tilt}) is therefore not an empirical improvement-rate calibration.

\paragraph{Progress retention under a KL budget.}
Let $\pi_W=\pi_{\mathrm{walk}}$ and assume the achieved full-sample KL on the same $\rho$ obeys
\begin{equation}
\mathbb{E}_{q\sim\rho}D_{\mathrm{KL}}\bigl(\pi_W(\cdot\mid q)\|\pi_R(\cdot\mid q)\bigr)\leq\kappa,\qquad
\varepsilon_\kappa=\min\{1,\sqrt{\kappa/2}\}.
\label{eq:app_theory_kl}
\end{equation}
Pinsker's inequality gives conditional total variation at most $\sqrt{D_{\mathrm{KL}}/2}$; Jensen's inequality then bounds its mean by $\varepsilon_\kappa$. Hence any $[0,1]$-valued statistic changes by at most $\varepsilon_\kappa$, including $P$ and event indicators. For $a_R=\Pr_R(P\geq p_0)$ and $f_W=\Pr_W(S<s_0)$,
\begin{equation}
\mathbb{E}_W[P]\geq\mathbb{E}_R[P]-\varepsilon_\kappa,\qquad
\Pr_W(P\geq p_0,S\geq s_0)\geq[a_R-\varepsilon_\kappa-f_W]_+.
\end{equation}
On the latter event, $SQ\geq\alpha p_0s_0$, so
\begin{equation}
J(\pi_W)\geq\alpha p_0s_0[a_R-\varepsilon_\kappa-f_W]_+.
\label{eq:app_theory_lower}
\end{equation}
The bound is nonvacuous only when $a_R>\varepsilon_\kappa+f_W$. A GRPO KL coefficient does not by itself establish Eq.~(\ref{eq:app_theory_kl}): the bound concerns the complete output distribution on $\rho$, not an unconverted average over tokens or diffusion steps. By data processing, full-output KL also bounds the induced trajectory KL. Conversely, reducing the probability of a fixed failure event by $d$ requires $\kappa\geq2d^2$ under this bound. Thus zero KL forbids repair, while a very large KL gives little progress retention.

\paragraph{When the final score increases.}
Safety improvement and progress can be correlated, so marginal submetric improvements are insufficient. Define $s_t=\mathbb{E}_t[S]>0$ and $c_t=\mathbb{E}_t[SQ]/s_t$ for $t\in\{R,W\}$. If the safety-weighted quality loss is bounded by $c_W\geq c_R-\ell$, then
\begin{equation}
\begin{aligned}
J(\pi_W)-J(\pi_R)
&=s_Wc_W-s_Rc_R\\
&=(s_W-s_R)c_R+s_W(c_W-c_R)\\
&\geq(s_W-s_R)c_R-s_W\ell.
\end{aligned}
\label{eq:app_theory_gain}
\end{equation}
Therefore $(s_W-s_R)c_R>s_W\ell$ is sufficient for strict improvement. This condition allows some scenes to worsen and can be estimated from paired per-trajectory scores; separate averages of NC, DAC, and EP do not prove it. If $s_R=0$, any positive Walk score is an improvement.

\paragraph{Sampling and boundary conditions.}
For a query with $p_q=\Pr_R(H=1,P\geq p_0\mid q)$, an independent group of $G$ samples contains a useful safe-progress candidate with probability $1-(1-p_q)^G$. When $p_q$ is small, the Walk signal is rarely observed; if all group members fail the gate, their planning rewards are all zero and there is no relative planning signal from that term. Finite KL cannot create probability on a reference-null event. These facts justify switching after progress discovery but before the Run checkpoint loses useful safe support, without implying a universal epoch or schedule.

Finally, $H$ is a training gate rather than the full evaluation factor $S$: it permits $\mathrm{NC}=1/2$, checks TTC, and omits EPDMS's DDC/TLC terms. Endpoint proximity similarly certifies neither path safety nor progress; a tight endpoint reward may increase $\ell$, while a loose one may be nondiscriminative. Human-penalty filters, Navhard's multi-stage aggregation, and policy-induced query shifts require protocol-specific analysis. These qualifications apply equally to autoregressive samples and VLM-conditioned diffusion trajectories.

\section{Training Hyperparameters}
\label{app:hyperparams}

Table~\ref{tab:hyperparams_full} summarizes the RL training hyperparameters for different VLM-based planners and stages. Note that we stop the first (Run) stage once EP has essentially plateaued, as illustrated in Fig.~\ref{fig:toy_run_walk}. This is why ReCogDrive requires a longer Run stage than AutoDrive-P$^3$. AutoDrive-P$^3$ starts from an SFT checkpoint with EP around 81.7, whereas ReCogDrive starts from an SFT checkpoint with EP around 80. Therefore, ReCogDrive requires a longer progress-discovery phase before EP plateaus, specifically two additional Run epochs compared with AutoDrive-P$^3$, using five Run epochs in total.

In the Run stage, both AutoDrive-P$^3$ and ReCogDrive adopt Eq.~\ref{eqa:run_reward} as the planning reward, i.e., PDMS is directly used as the planning reward to enhance progress exploration. In the Walk stage, both planners adopt Eq.~\ref{eqa:walk_reward} as the planning reward to improve safety. The difference is that AutoDrive-P$^3$ additionally includes a planning/perception auxiliary reward and combines it with the planning reward through Eq.~\ref{eqa:mix_reward}, where \(R_{\mathrm{aux}}(x_i)\) denotes the auxiliary reward, to align with its original baseline \citep{ye2026autodrive}. ReCogDrive does not require this additional auxiliary term, as it is consistent with its original baseline \citep{li2025recogdrive}.

\begin{table}[t]
\centering
\caption{\textbf{RL training hyperparameters for different VLM-based planners.}}
\label{tab:hyperparams_full}
\resizebox{1.0\linewidth}{!}{
\begin{tabular}{lcccc}
\toprule
\multirow{2}{*}{\textbf{Hyperparameter}}  & \multicolumn{2}{c}{\textbf{AutoDrive-P$^3$ \citep{ye2026autodrive}}} & \multicolumn{2}{c}{\textbf{ReCogDrive\citep{li2025recogdrive}}} \\
\cmidrule(lr){2-3}\cmidrule(lr){4-5}
 & \textbf{Run-GRPO} & \textbf{Walk-GRPO} & \textbf{Run-GRPO} & \textbf{Walk-GRPO} \\
\midrule
Learning Rate & $1\times10^{-6}$ & $1\times10^{-6}$ & $1\times10^{-4}$ & $1\times10^{-6}$ \\
LR Schedule & constant & constant & cosine decay & cosine decay \\
Optimizer & AdamW & AdamW & AdamW & AdamW \\
RL Epochs & 3 & 2 & 5 & 1 \\
Global Batch Size & 32 & 32 & 32 & 32 \\
GRPO Group Size ($G$) & 8 & 8 & 8 & 8 \\
Clip Parameter ($\epsilon$) & 0.2 & 0.2 & 0.2 & 0.2 \\
Precision / Hardware & bf16 / $8\times$A100 & bf16 / $8\times$A100 & bf16 / $8\times$A100 & bf16 / $8\times$A100 \\
Trajectory Output & 8 waypoints (4.0 s) & 8 waypoints (4.0 s) & 8 poses (4.0 s, 0.5 s) & 8 poses (4.0 s, 0.5 s) \\
Planner Parameterization & Autoregressive tokens & Autoregressive tokens & Diffusion & Diffusion \\
\bottomrule
\end{tabular}}
\end{table}

\section{Stage-by-Stage Qualitative Walkthrough: SFT $\rightarrow$ Run $\rightarrow$ Walk}
\label{app:stage_qualitative}

\begin{figure}[t]
\centering
\begin{subfigure}{0.78\linewidth}
\includegraphics[width=\linewidth]{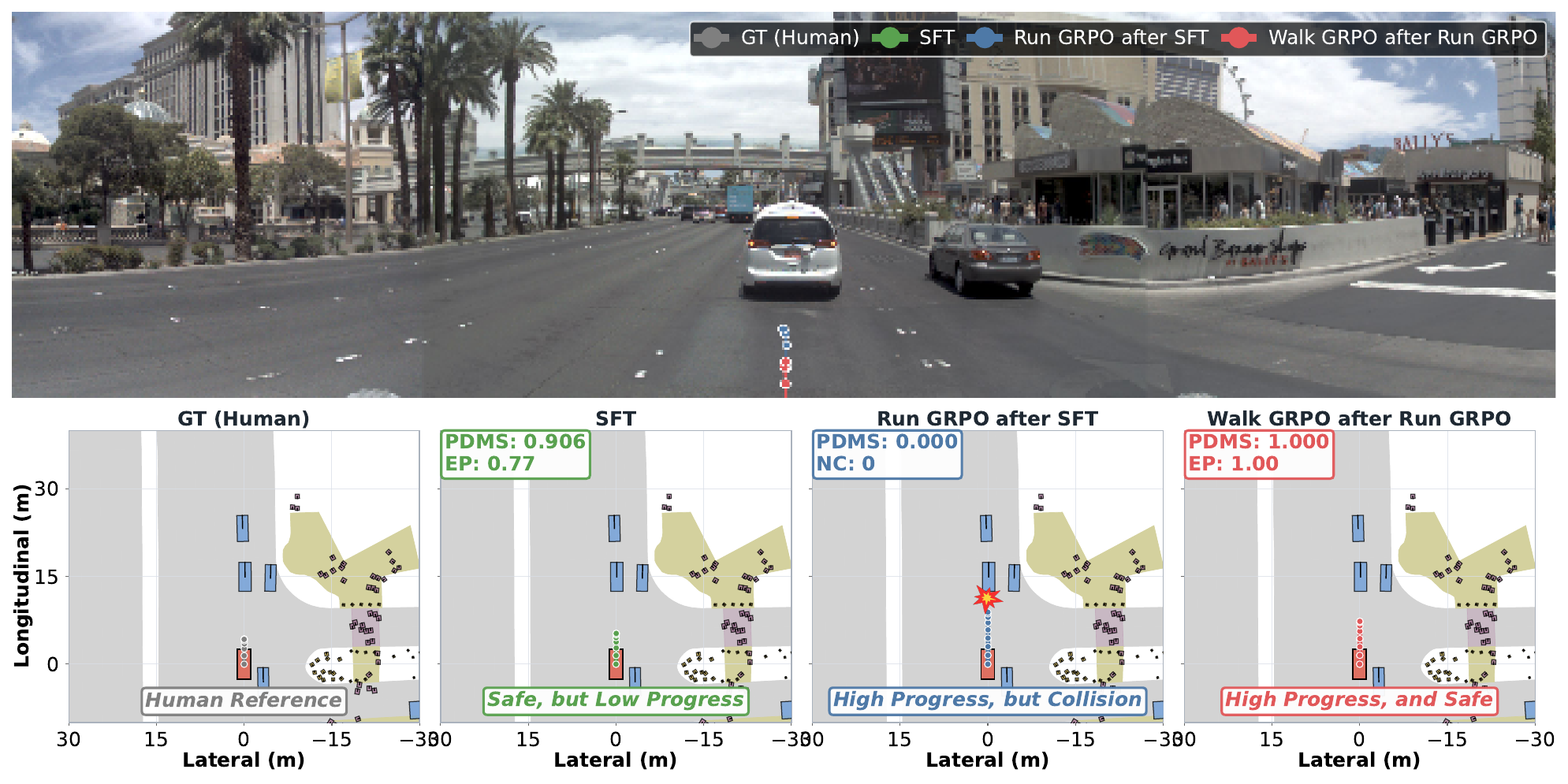}
\caption{Straight urban cruising: SFT stops short (EP\,$0.77$); Walk reaches full progress (EP\,$1.00$, PDMS\,$1.00$) while staying centered in lane.}
\end{subfigure}

\vspace{2.5mm}
\begin{subfigure}{0.78\linewidth}
\includegraphics[width=\linewidth]{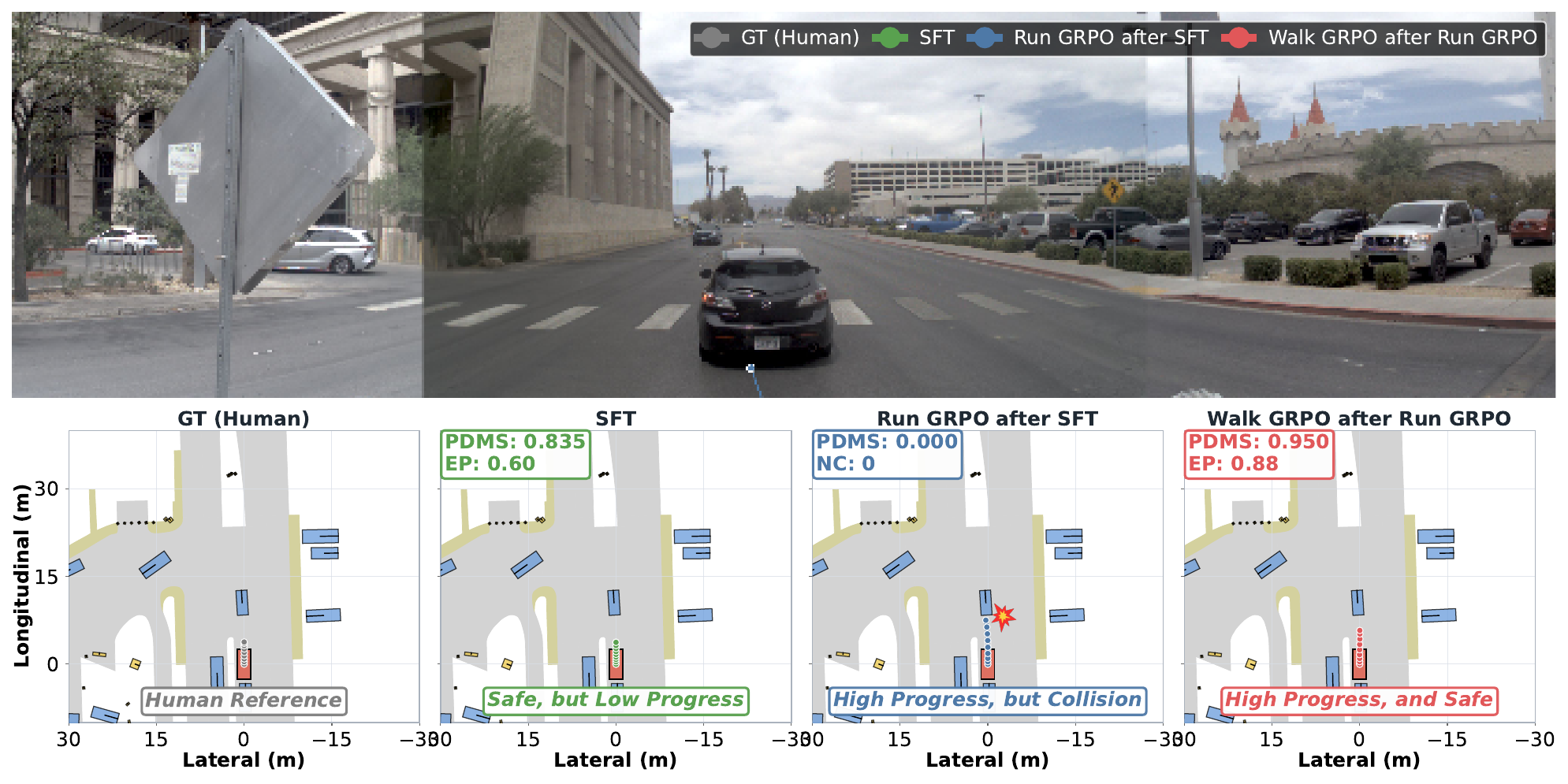}
\caption{Open road: the Run policy overshoots off-road (PDMS\,$0$); Walk restores a safe path that still advances far (EP $0.60\!\rightarrow\!0.88$, PDMS $0.835\!\rightarrow\!0.950$).}
\end{subfigure}

\vspace{2.5mm}
\begin{subfigure}{0.78\linewidth}
\includegraphics[width=\linewidth]{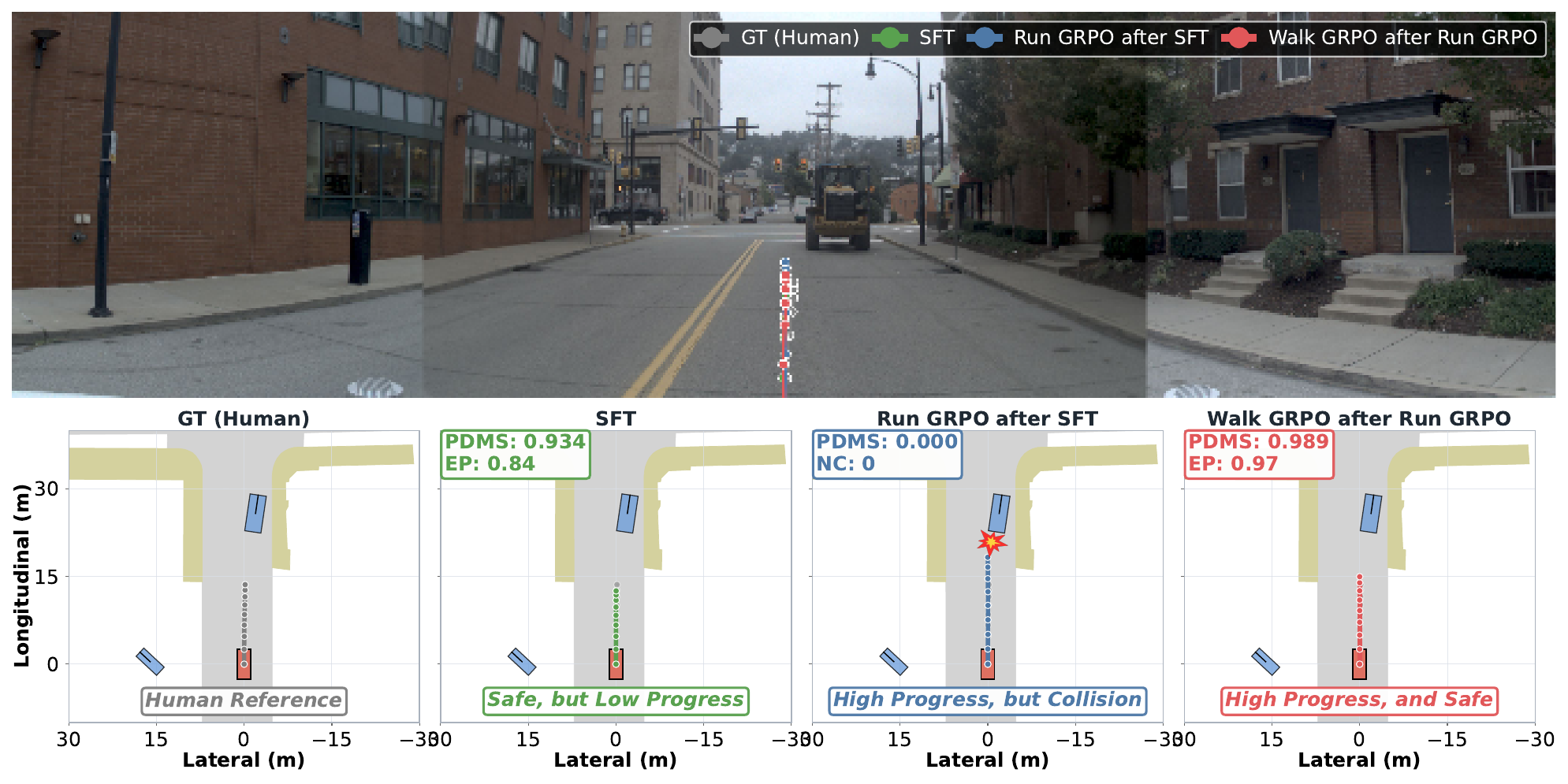}
\caption{Intersection turn: Walk tracks the human arc more tightly than SFT while extending progress (EP $0.84\!\rightarrow\!0.97$, PDMS $0.934\!\rightarrow\!0.989$).}
\end{subfigure}
\caption{\textbf{Stage-by-stage qualitative results (part 1/2).} Human\,/\,SFT\,/\,Run\,/\,Walk trajectories on three NAVSIM scenes. SFT is safe but short, Run is far but unsafe (NC\,$=0$), and Walk is both far and safe.}
\label{fig:stage_cases_a}
\end{figure}

\begin{figure}[t]
\centering
\begin{subfigure}{0.78\linewidth}
\includegraphics[width=\linewidth]{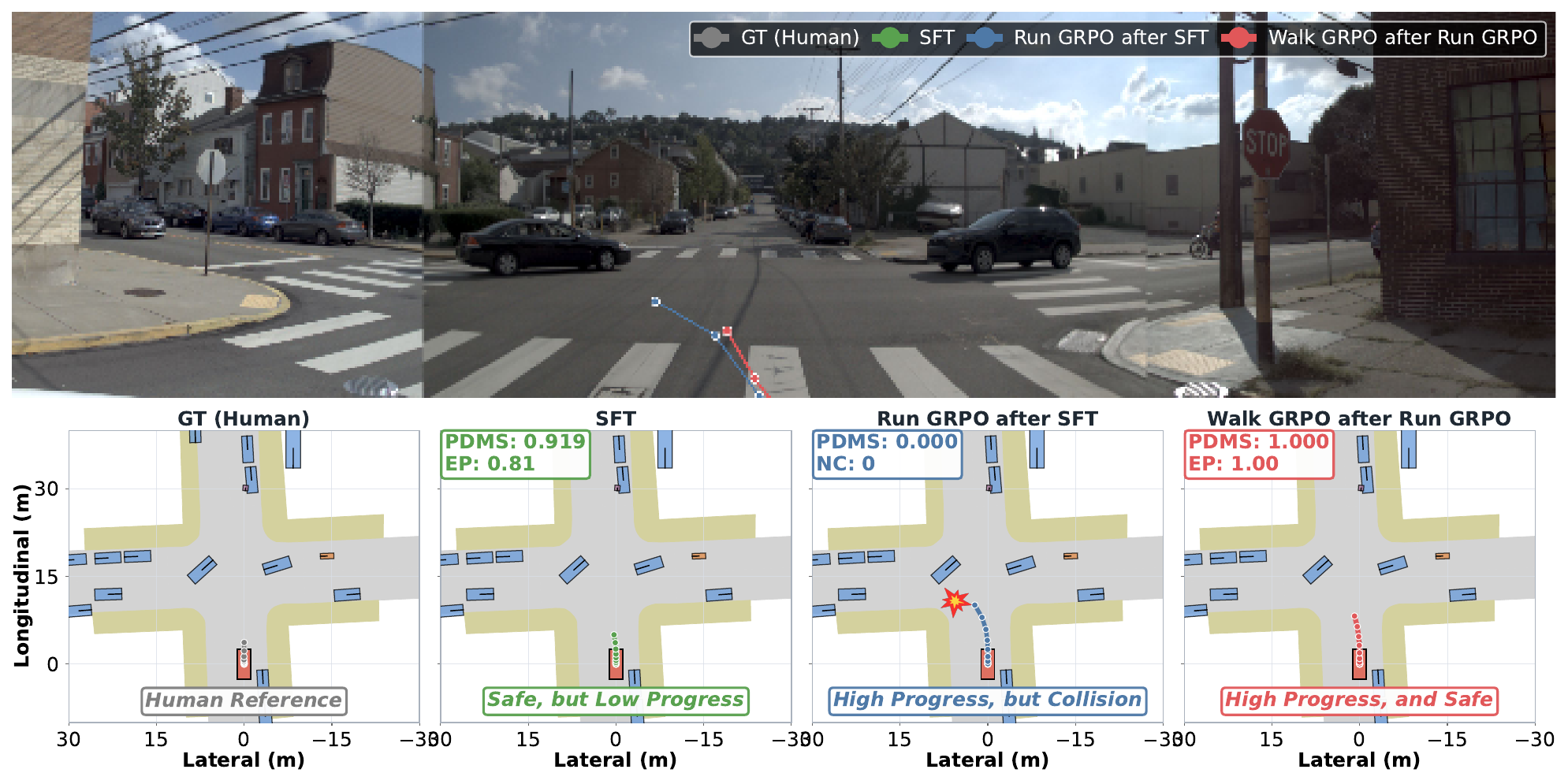}
\caption{Crosswalk approach: Walk yields safely yet still clears the junction (EP $0.81\!\rightarrow\!1.00$, PDMS $0.919\!\rightarrow\!1.00$).}
\end{subfigure}

\vspace{2.5mm}
\begin{subfigure}{0.78\linewidth}
\includegraphics[width=\linewidth]{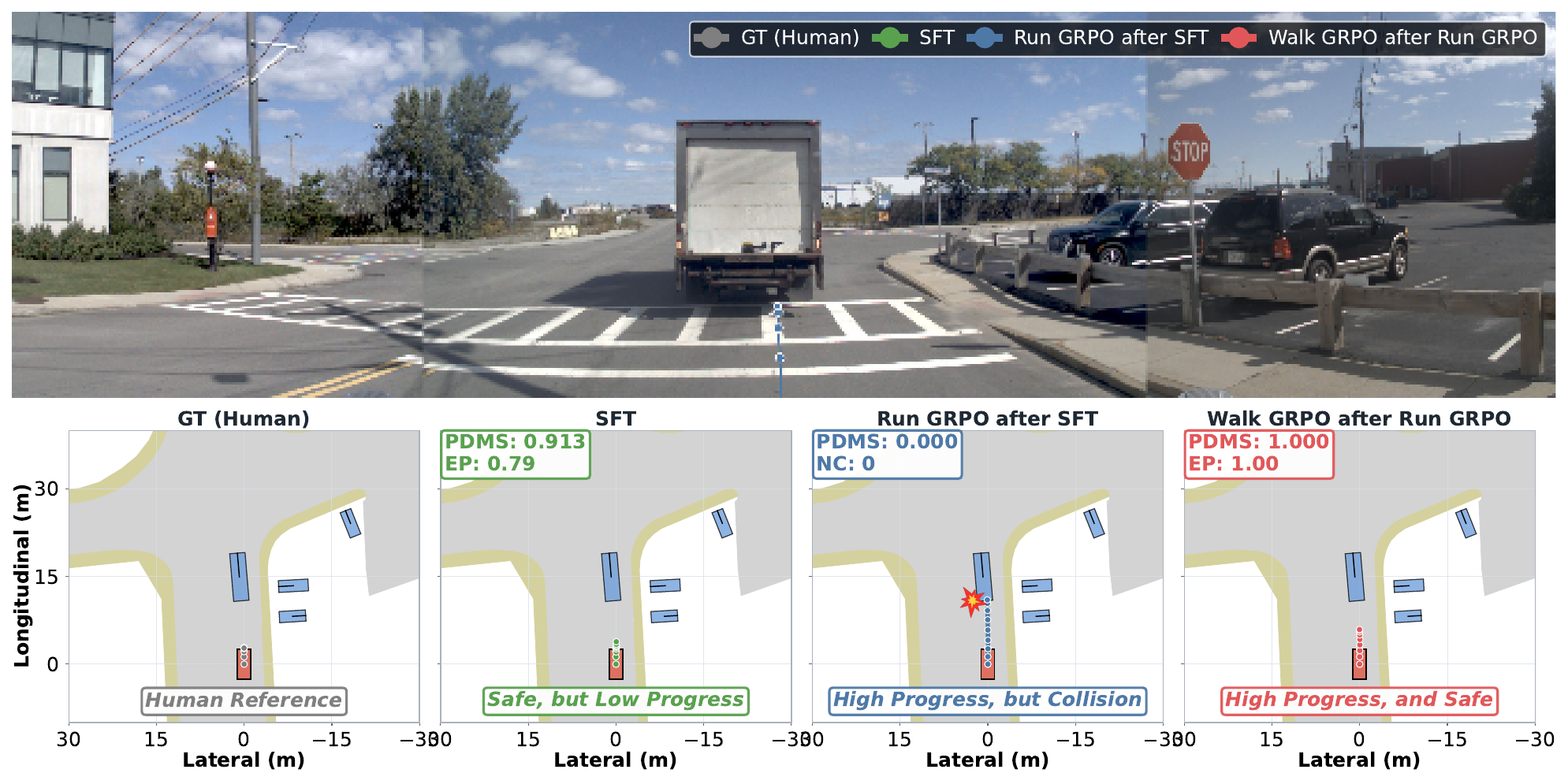}
\caption{Following a truck: SFT over-brakes behind the lead vehicle; Walk keeps a safe gap while progressing fully (EP $0.79\!\rightarrow\!1.00$, PDMS $0.913\!\rightarrow\!1.00$).}
\end{subfigure}

\vspace{2.5mm}
\begin{subfigure}{0.78\linewidth}
\includegraphics[width=\linewidth]{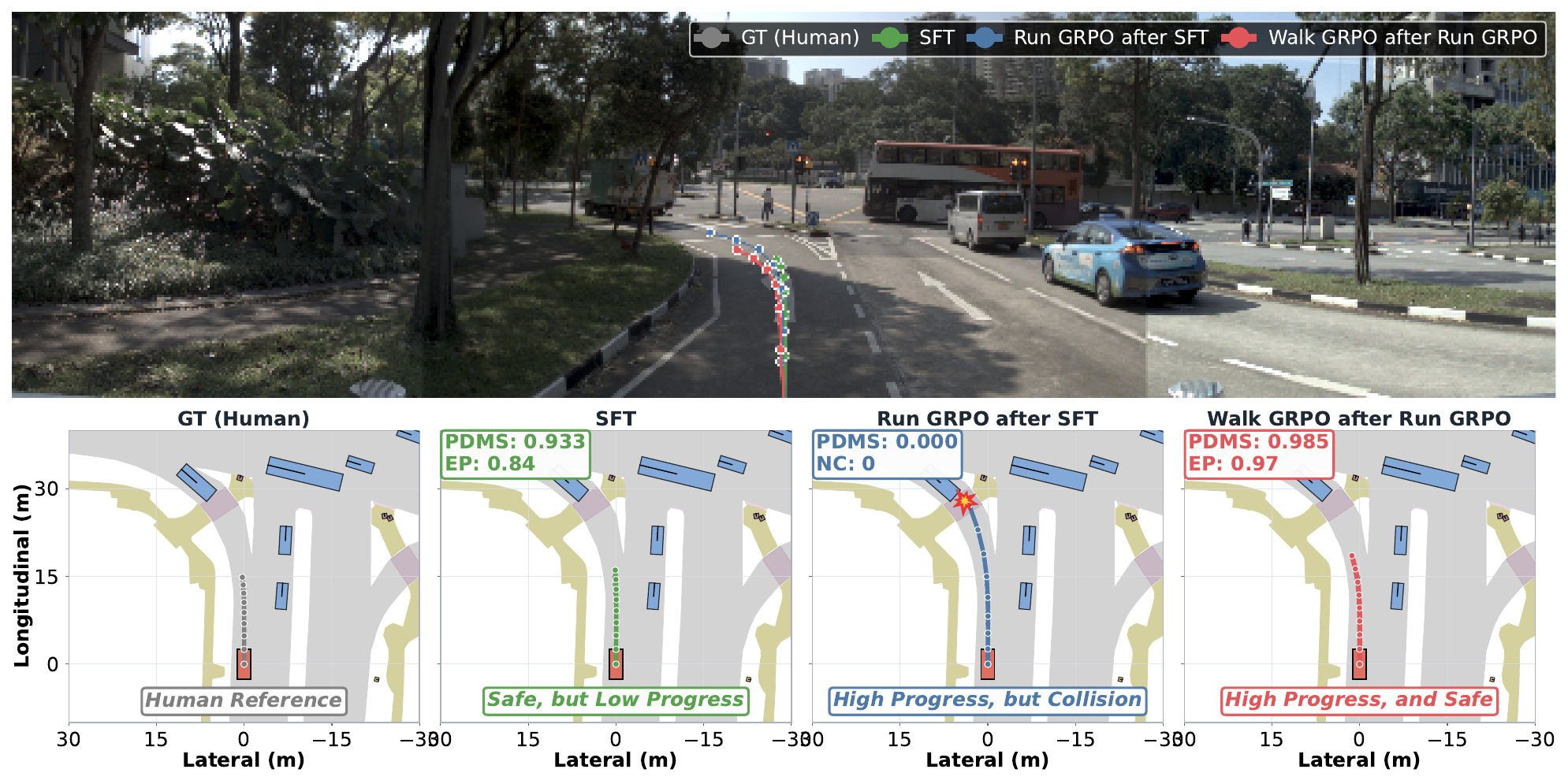}
\caption{Bus interaction: Walk advances past the SFT stopping point without violating safety (EP $0.84\!\rightarrow\!0.97$, PDMS $0.933\!\rightarrow\!0.985$).}
\end{subfigure}
\caption{\textbf{Stage-by-stage qualitative results (part 2/2).} Three additional scenes. In every case the Walk-GRPO policy (orange) inherits the long-range progress discovered by Run-GRPO (red) but removes its collisions/off-road events, ending close to or above the human progress level while remaining collision-free.}
\label{fig:stage_cases_b}
\end{figure}

This section visualizes how a single scene evolves through the three training stages of our pipeline. In Figures~\ref{fig:stage_cases_a} and~\ref{fig:stage_cases_b}, each row contains a wide camera strip (with all four trajectories projected) followed by four bird's-eye-view (BEV) panels that share an identical map/agent layout. From left to right the BEV panels show the \emph{Human} expert (GT, green), the \emph{SFT} policy (blue), the \emph{Run-GRPO} checkpoint (red), and the final \emph{Walk-GRPO} policy (orange). The PDMS and endpoint-progress (EP) scores are annotated inside every panel.

\textbf{Stage 1: SFT (conservative imitation).}
The supervised policy faithfully imitates the human demonstrations and is therefore collision-free, but it is systematically \emph{under-progressing}: it brakes early and stops short of where a competent human driver would be, yielding low EP and a capped PDMS (typically $0.83$--$0.93$ in these scenes). The blue BEV panel is labeled \emph{``Safe, but Low Progress.''}

\textbf{Stage 2: Run-GRPO (progress discovery).}
Optimizing the PDMS-only planning reward without any endpoint/safety constraint pushes the policy to discover much more aggressive, high-progress trajectory modes. As the red BEV panel (\emph{``High Progress, but Collision''}) shows, the Run policy reaches far further down the road, but it frequently leaves the drivable area or collides with an agent, driving its no-collision (NC) score, and hence PDMS, to zero. This is the intentional, temporary safety regression that the Run phase trades for exploration.

\textbf{Stage 3: Walk-GRPO (safety repair).}
Re-introducing the endpoint and safety rewards \emph{after} the Run phase lets the policy keep the newly discovered progress while repairing the unsafe behavior. The orange BEV panel (\emph{``High Progress, and Safe''}) recovers a collision-free trajectory that nonetheless travels almost as far as the Run policy, lifting both EP and PDMS well above the SFT baseline (often to $0.95$--$1.00$ PDMS). Across the six scenes below, spanning straight cruising, signalized intersections, crosswalk approaches, and following large vehicles (buses/trucks), the Walk policy consistently dominates SFT on progress while matching its safety, which is precisely the behavior our Run-then-Walk ordering is designed to produce.

\section{Qualitative Comparison}
\label{app:baseline_qualitative}

\begin{figure}[p]
\centering
\begin{subfigure}{0.85\linewidth}
\includegraphics[width=\linewidth]{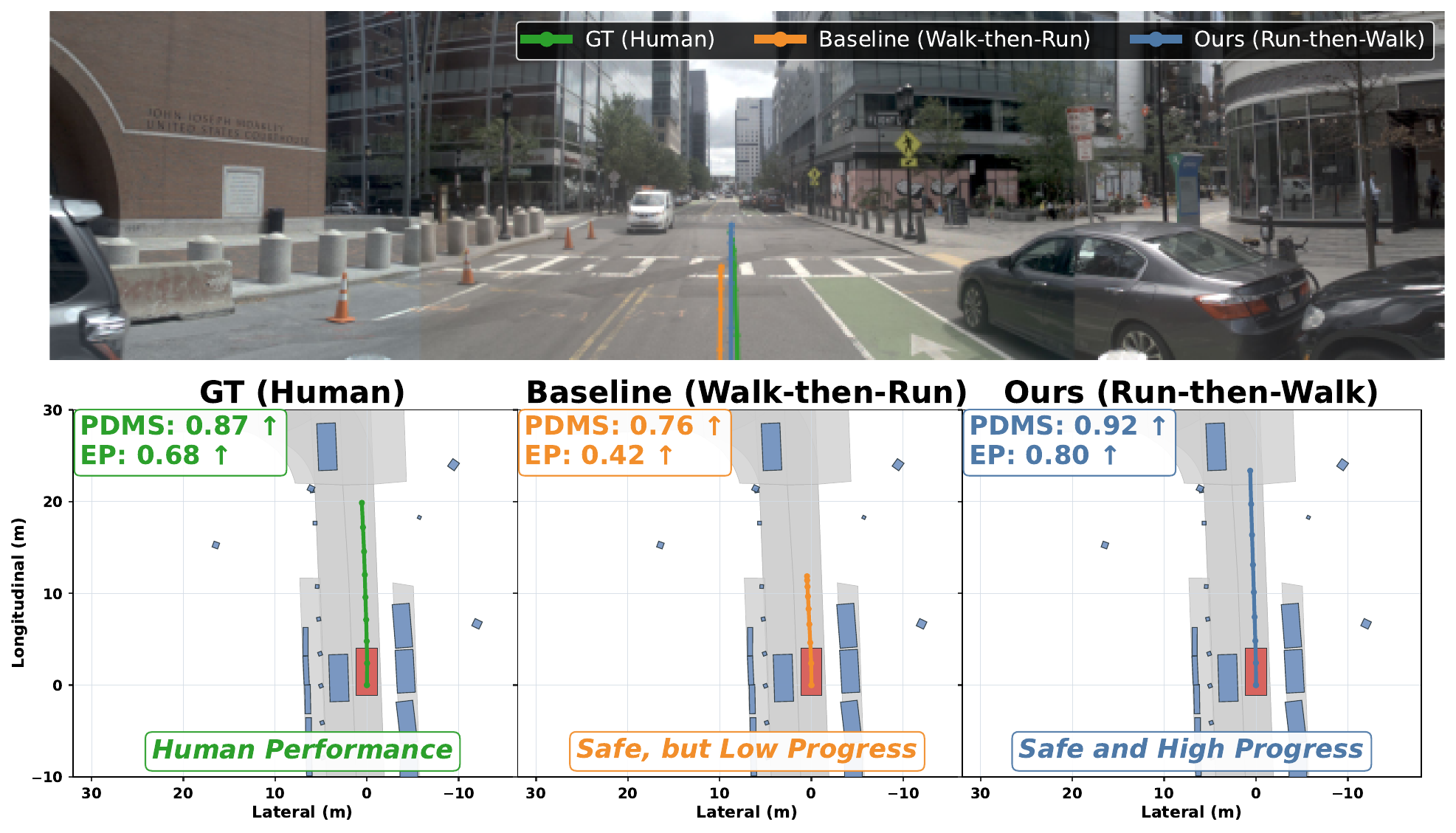}
\end{subfigure}

\begin{subfigure}{0.85\linewidth}
\includegraphics[width=\linewidth]{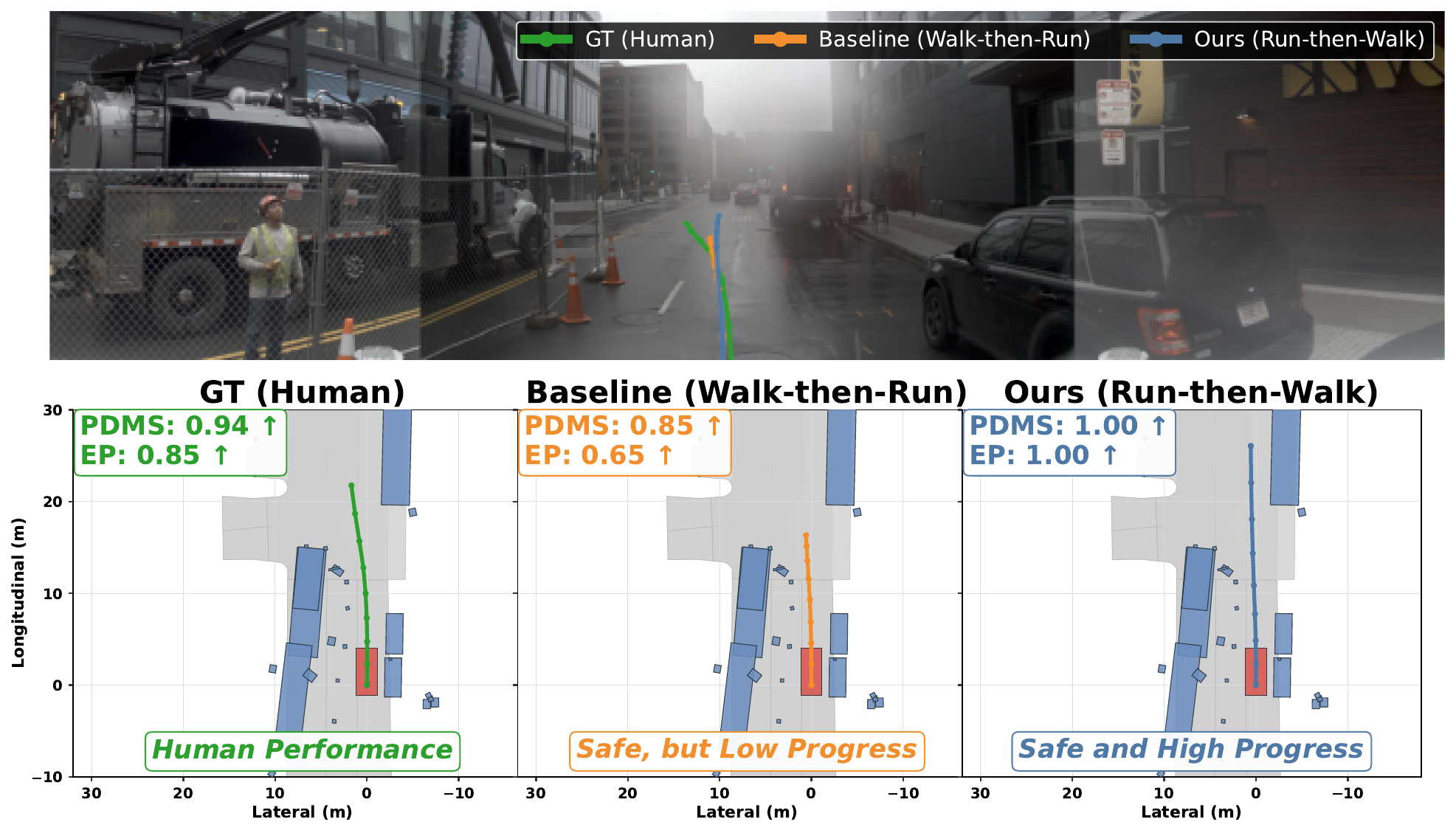}
\end{subfigure}

\begin{subfigure}{0.85\linewidth}
\includegraphics[width=\linewidth]{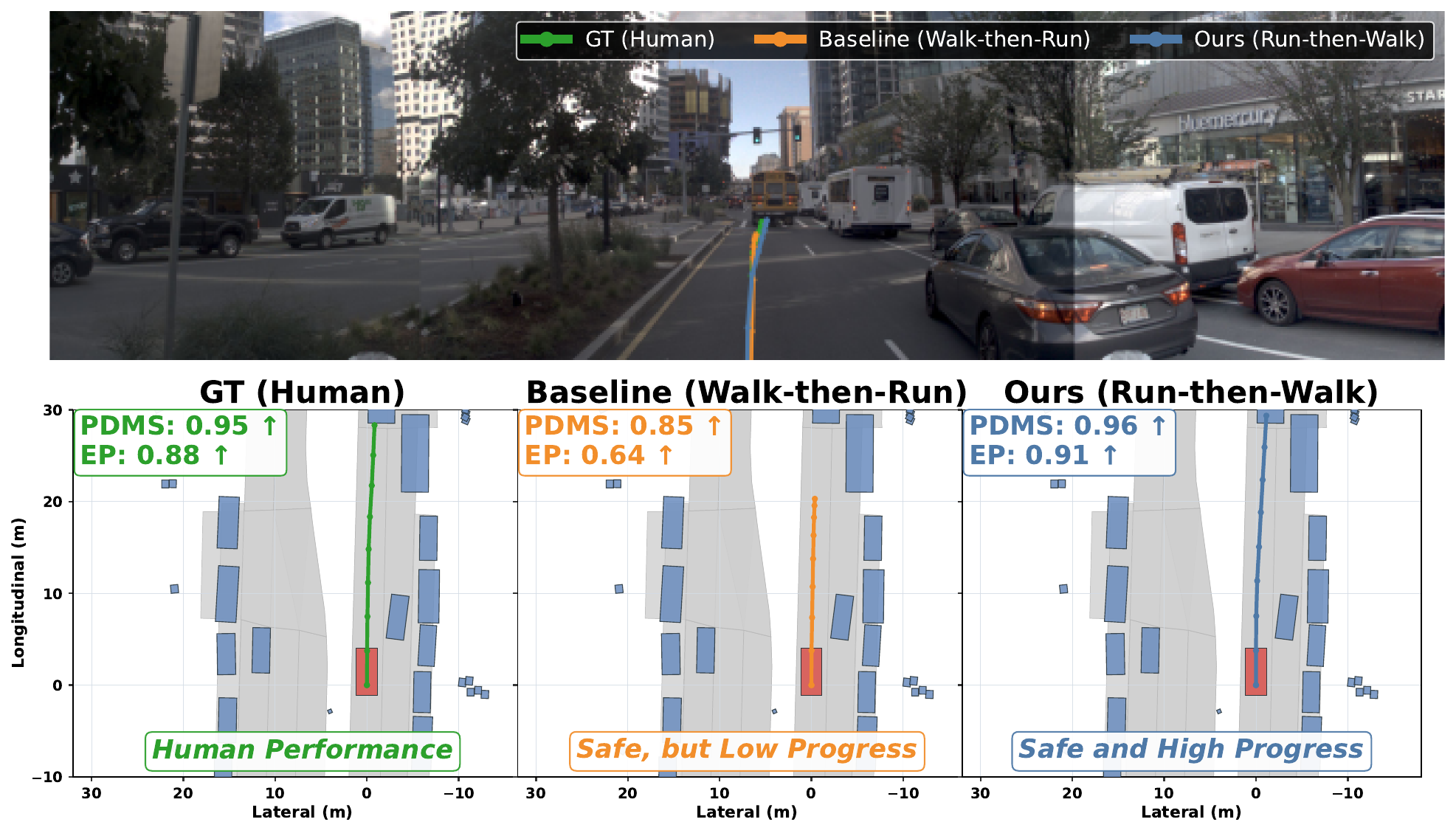}
\end{subfigure}
\caption{\textbf{Qualitative comparison with  Walk-then-Run (AutoDrive-P$^3$) part 1/2.} Across three scenes, the orange safety-first baseline stops short despite remaining safe. Run-then-Walk (blue) travels farther safely and raises both ego progress and PDMS, exposing the driving-efficiency cost of excessive conservatism.}
\label{fig:baseline_cases_b}
\end{figure}

\begin{figure}[p]
\centering
\begin{subfigure}{0.85\linewidth}
\includegraphics[width=\linewidth]{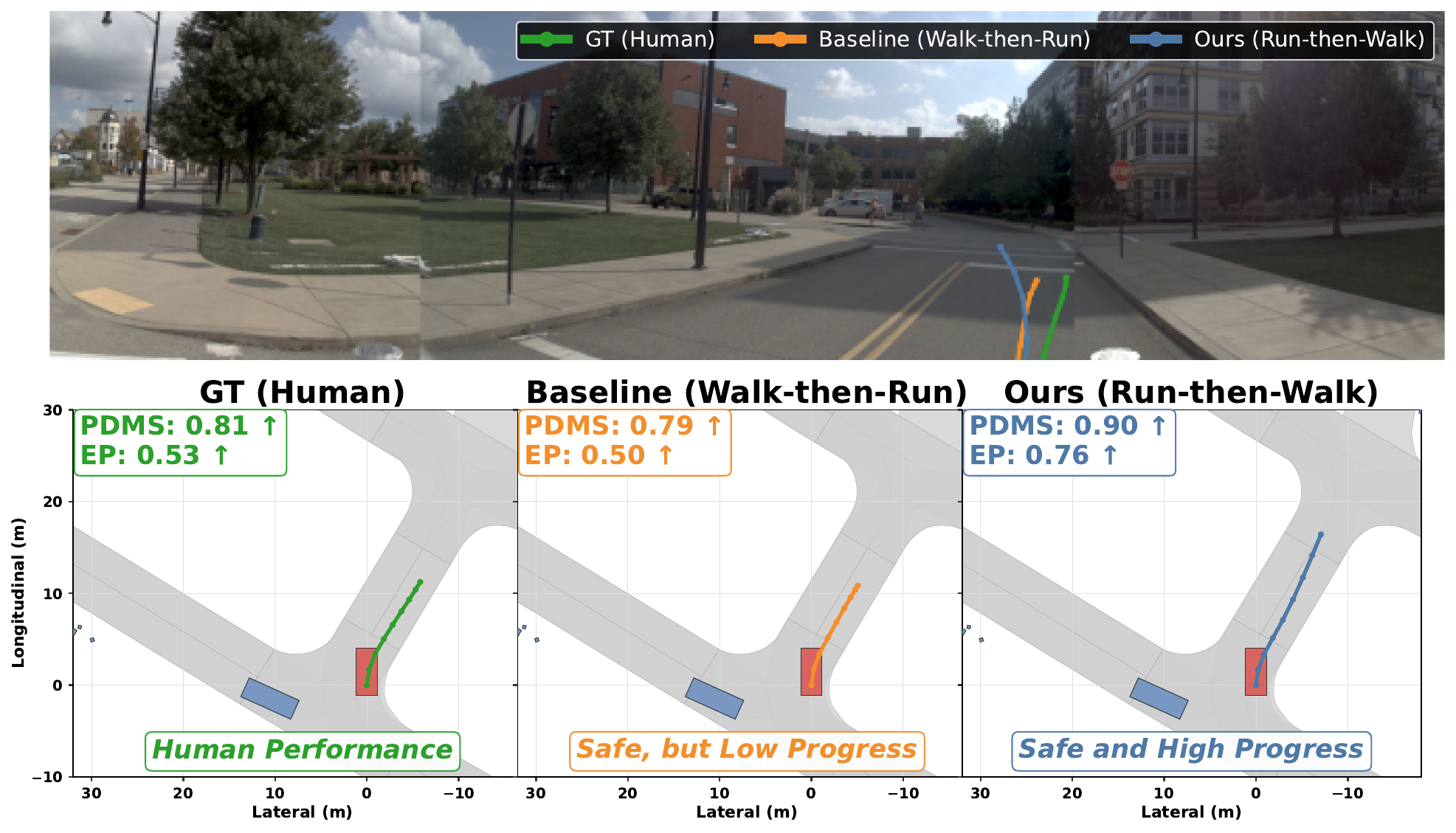}
\end{subfigure}

\begin{subfigure}{0.85\linewidth}
\includegraphics[width=\linewidth]{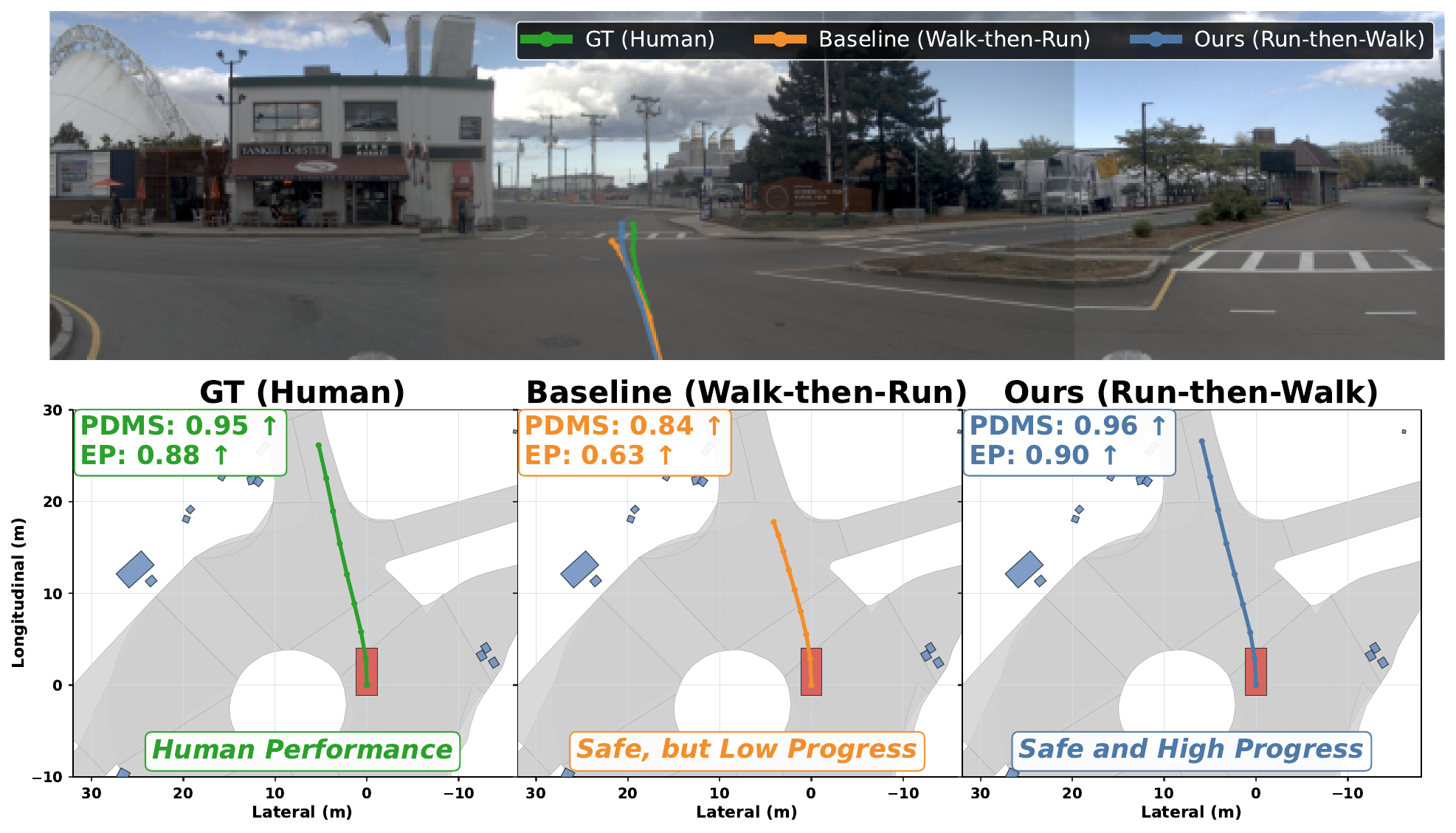}
\end{subfigure}

\begin{subfigure}{0.85\linewidth}
\includegraphics[width=\linewidth]{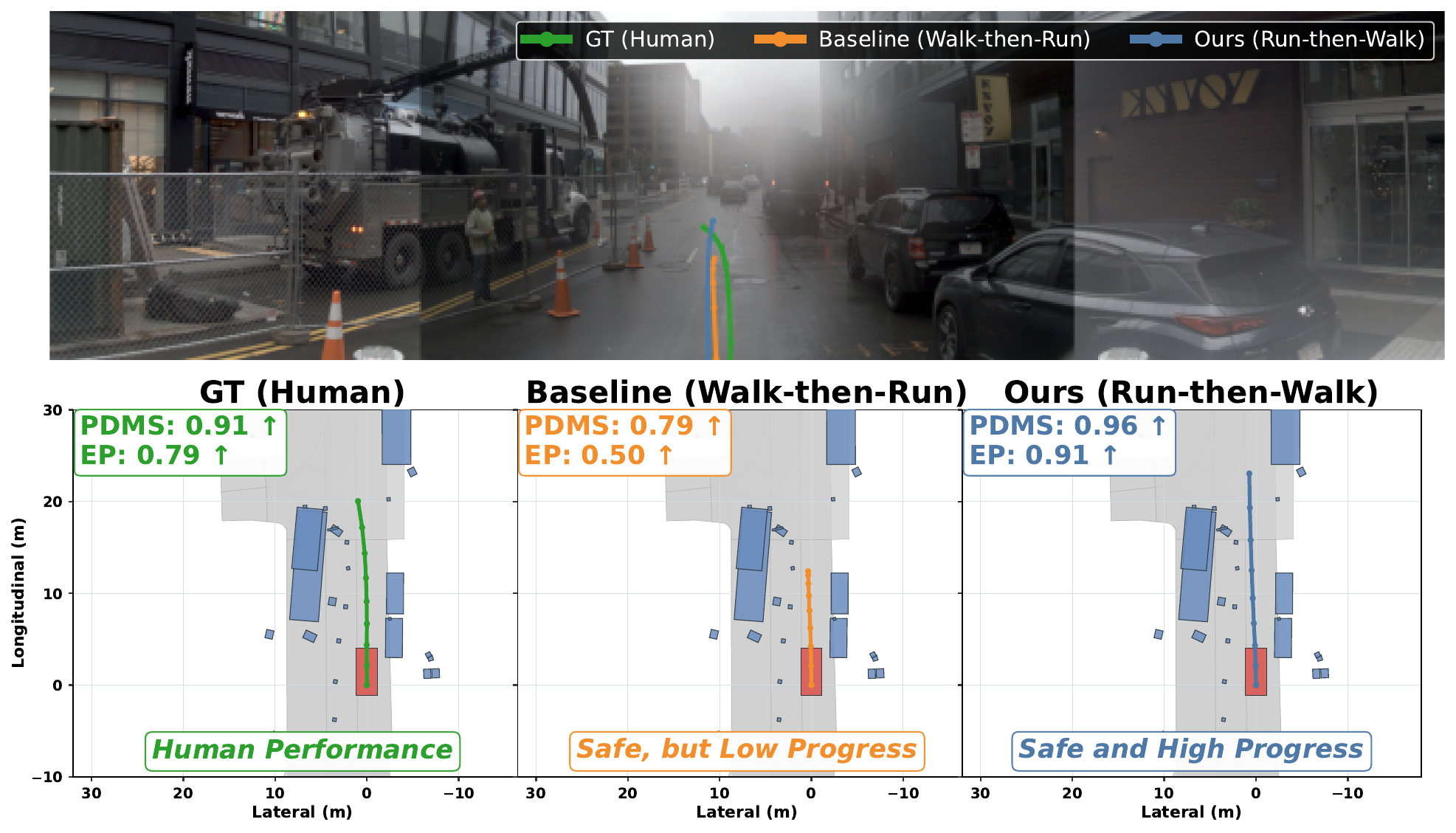}
\end{subfigure}
\caption{\textbf{Qualitative comparison with  Walk-then-Run (AutoDrive-P$^3$) part 2/2.} Three additional scenes confirm that the safety-first baseline remains safe but sacrifices endpoint progress. Run-then-Walk reaches farther safe endpoints and improves driving efficiency.}
\label{fig:baseline_cases_b_part2}
\end{figure}

\begin{figure}[p]
\centering
\begin{subfigure}{0.85\linewidth}
\includegraphics[width=\linewidth]{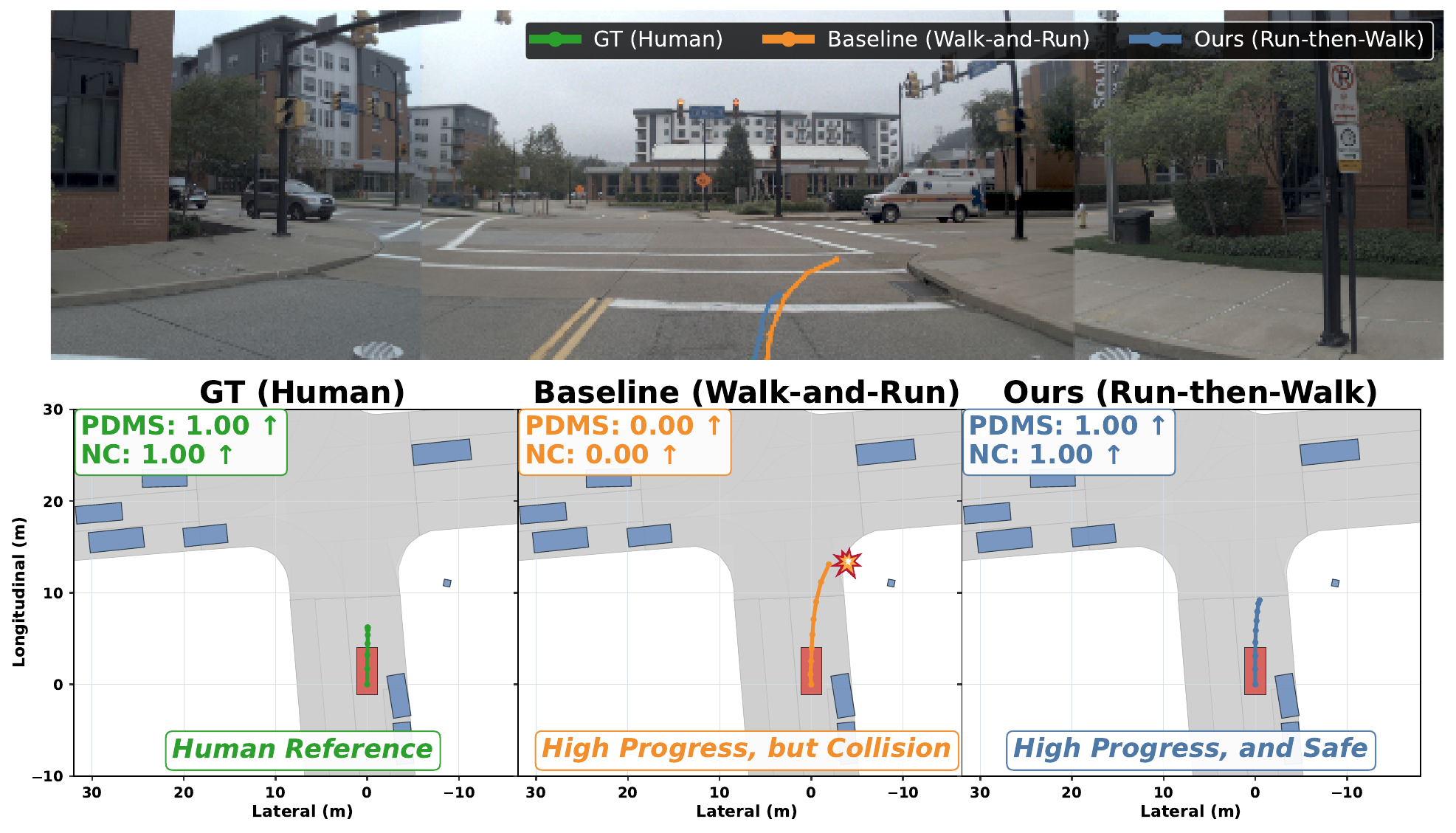}
\end{subfigure}

\begin{subfigure}{0.85\linewidth}
\includegraphics[width=\linewidth]{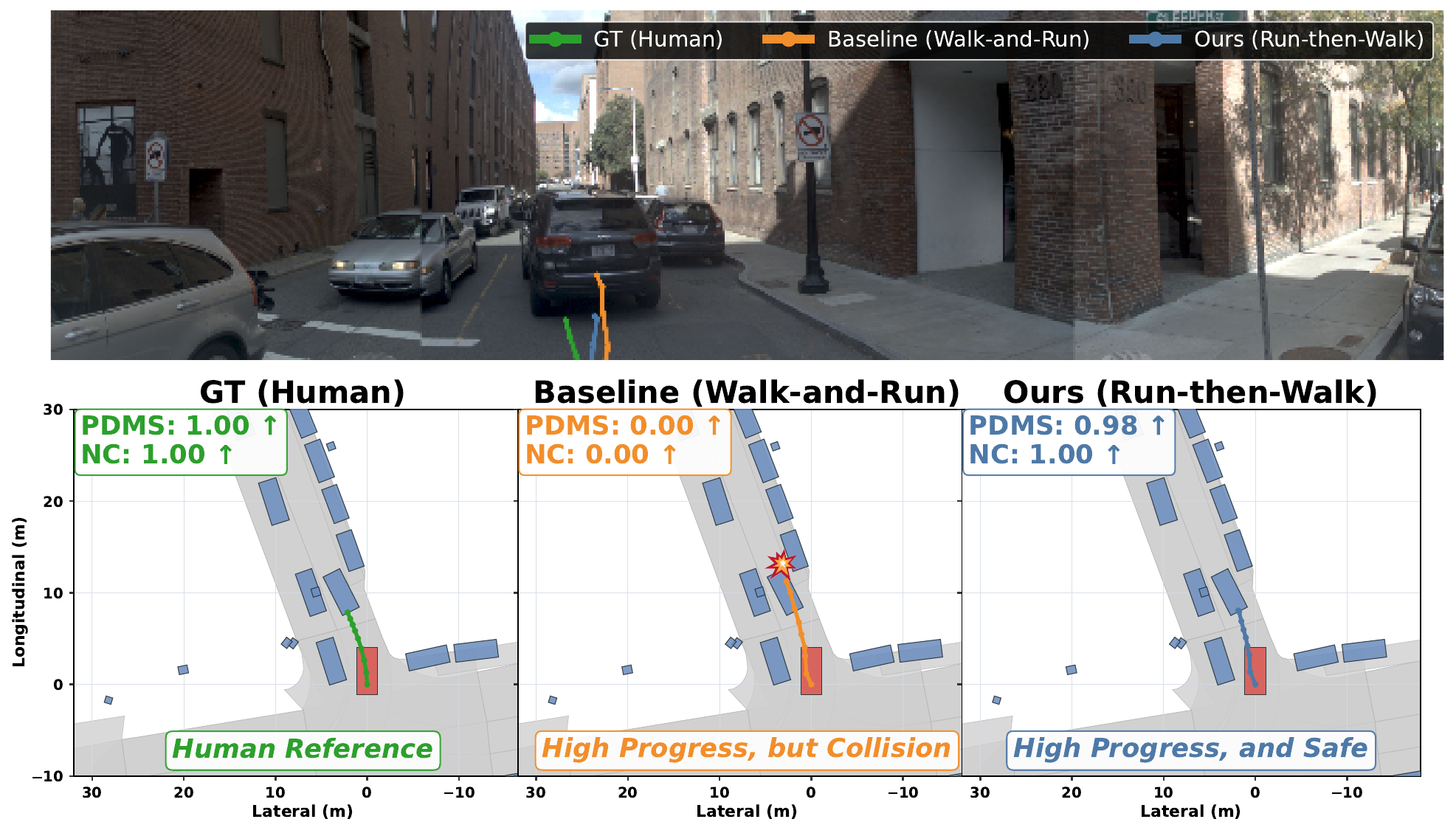}
\end{subfigure}

\begin{subfigure}{0.85\linewidth}
\includegraphics[width=\linewidth]{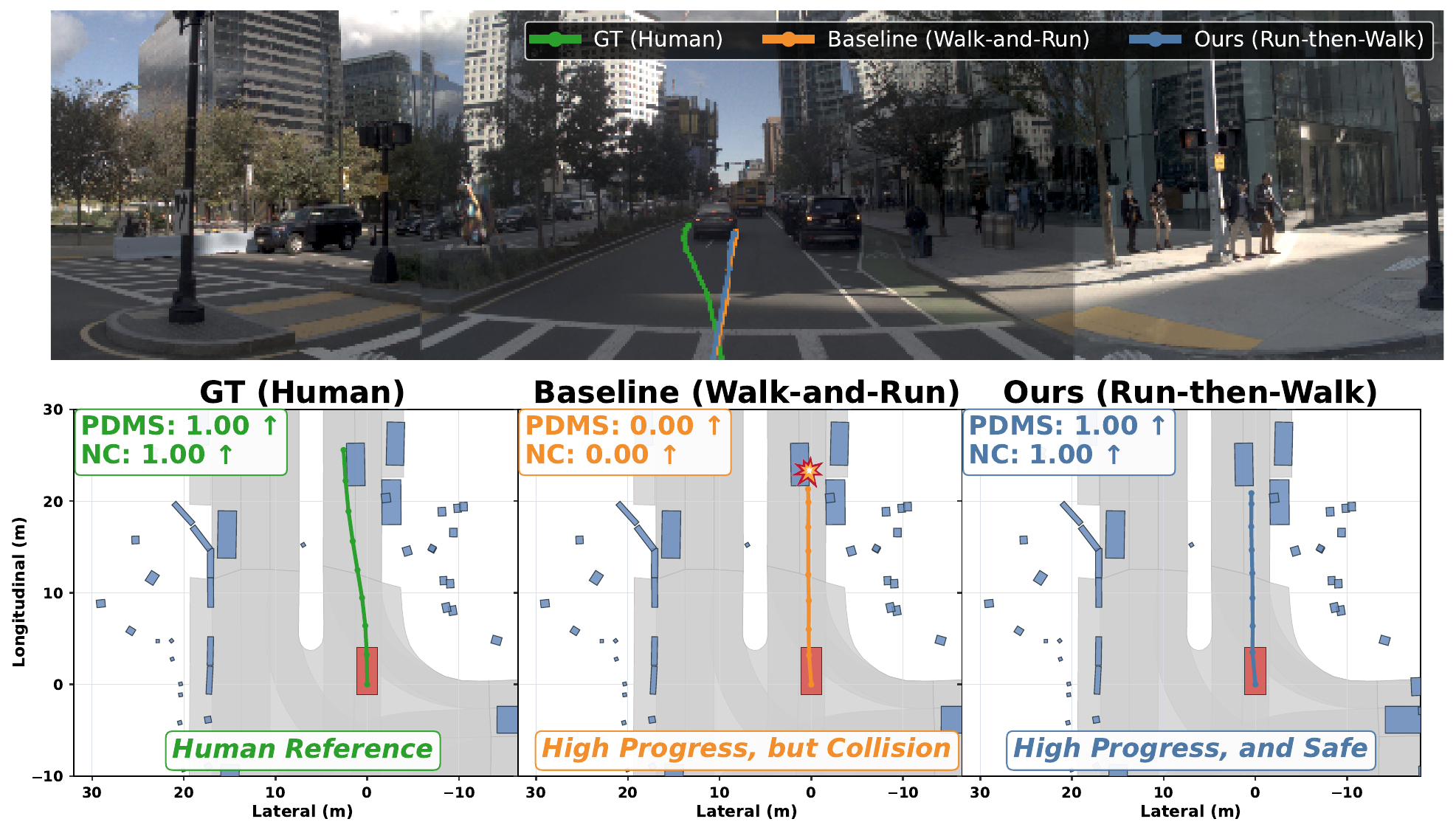}
\end{subfigure}
\caption{\textbf{Qualitative comparison with Walk-and-Run (ReCogDrive), part 1/2.} Three scenes show the cost of overemphasizing progress: the orange baseline advances aggressively but collides in each case. The blue Run-then-Walk trajectories continue forward without collision, avoiding the baseline's safety failure.}
\label{fig:baseline_cases_a}
\end{figure}

\begin{figure}[p]
\centering
\begin{subfigure}{0.85\linewidth}
\includegraphics[width=\linewidth]{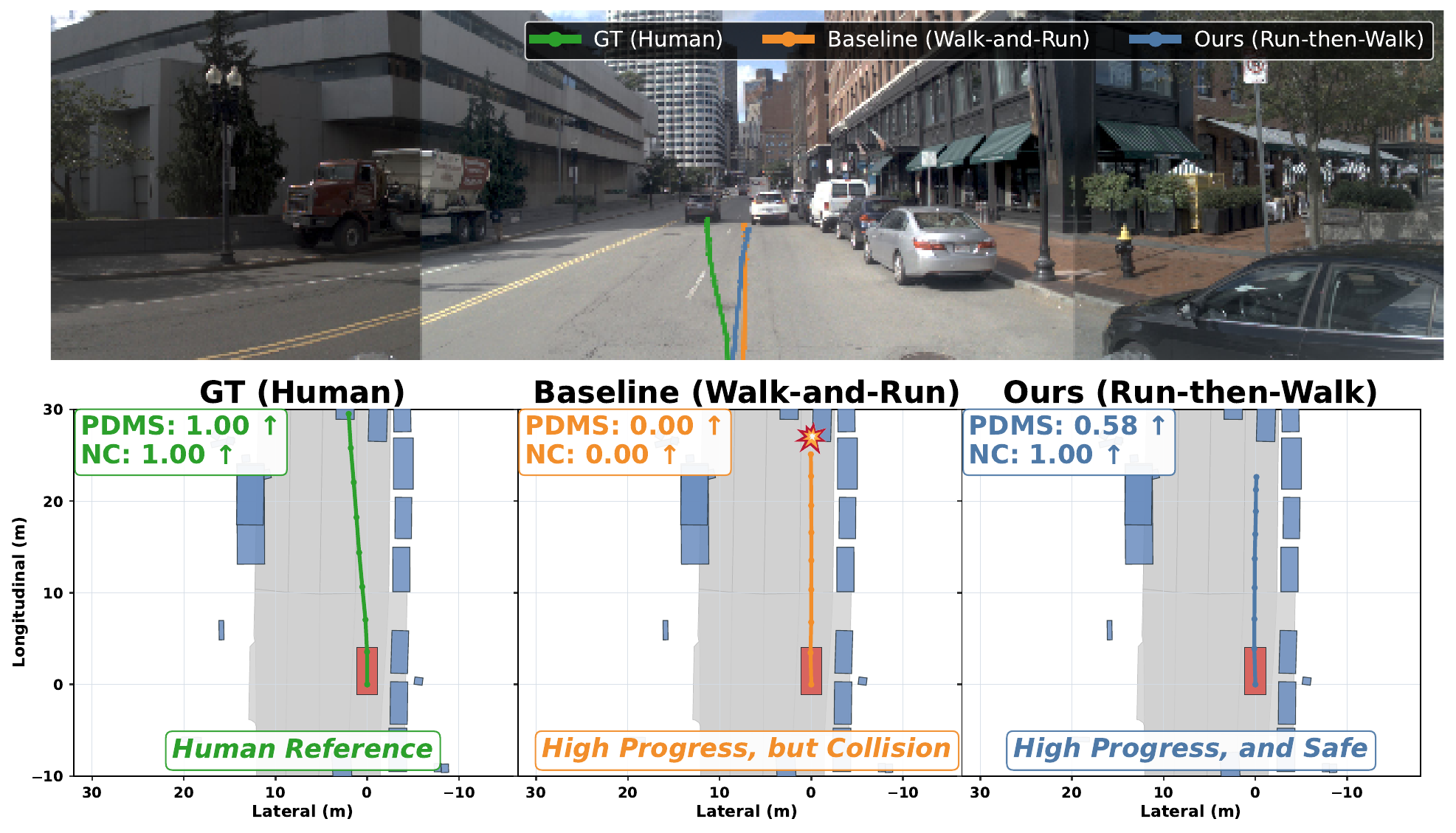}
\end{subfigure}

\begin{subfigure}{0.85\linewidth}
\includegraphics[width=\linewidth]{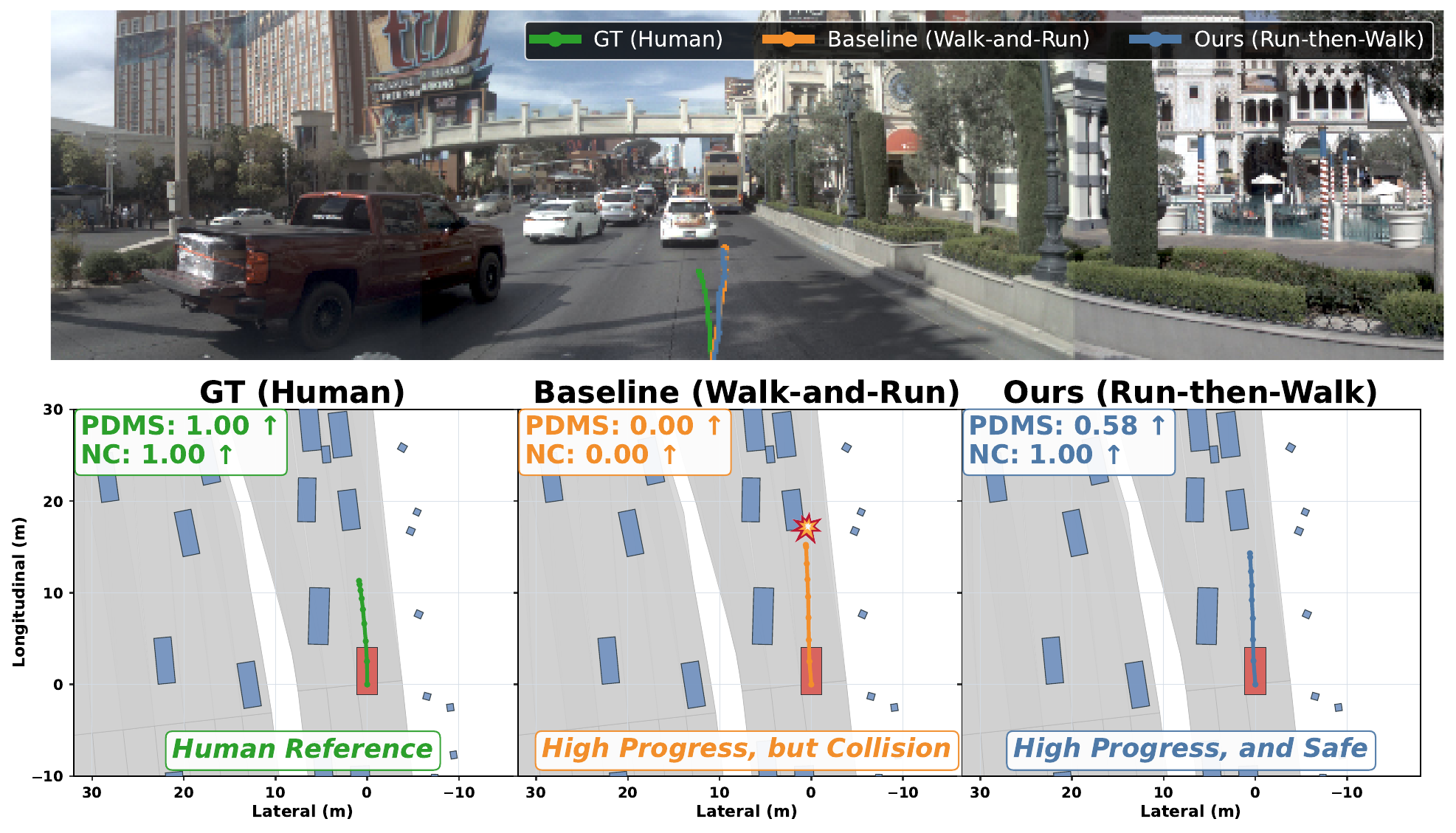}
\end{subfigure}

\begin{subfigure}{0.85\linewidth}
\includegraphics[width=\linewidth]{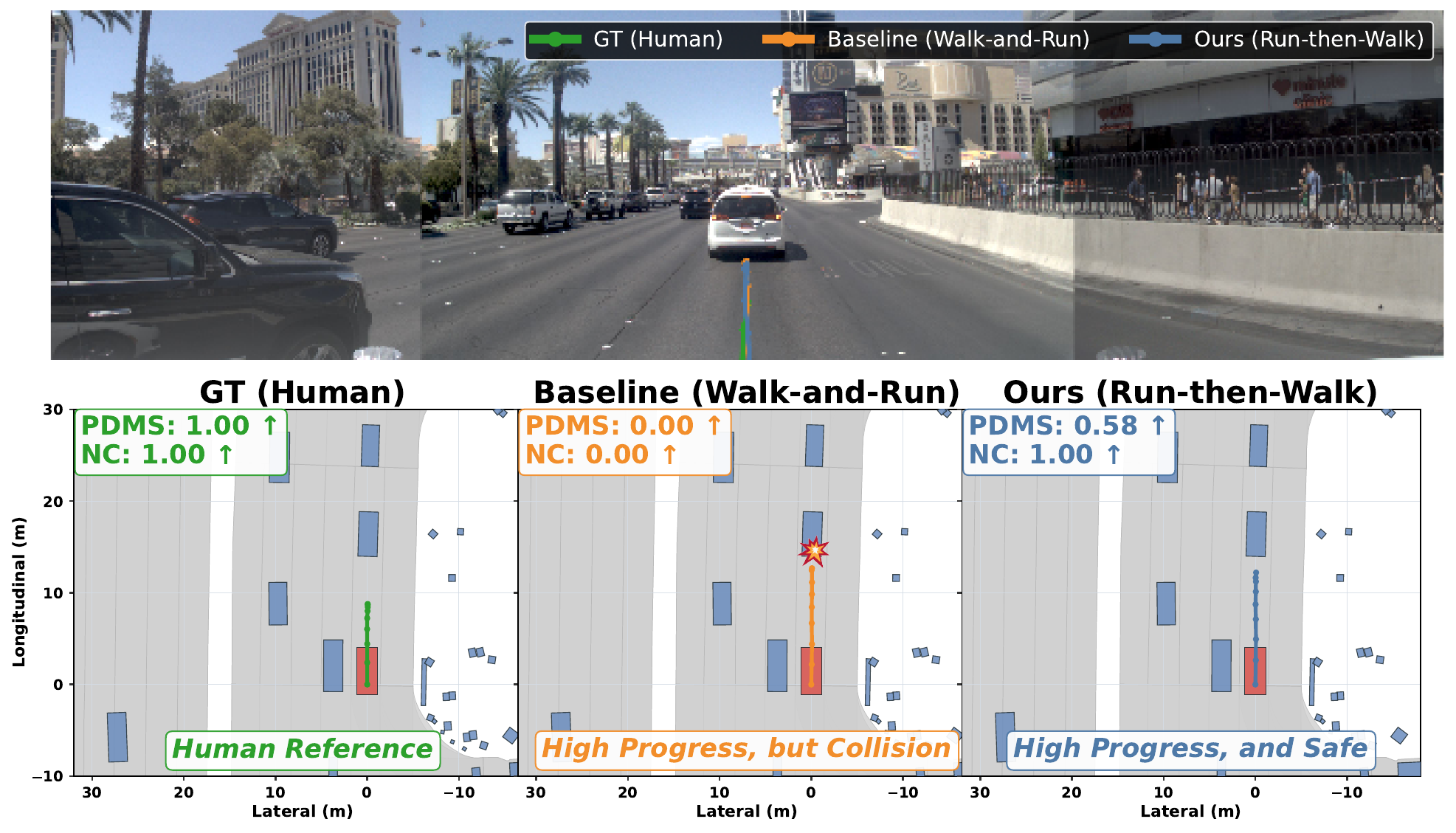}
\end{subfigure}
\caption{\textbf{Qualitative comparison with Walk-and-Run (ReCogDrive), part 2/2.} Three additional scenes show the same failure mode: progress-dominant training produces unsafe trajectories, while Run-then-Walk preserves forward motion with NC $1$.}
\label{fig:baseline_cases_a_part2}
\end{figure}

The following four pages present NAVSIM cases for a qualitative comparison. Each triplet shows the human reference (green), the corresponding baseline (orange), and Run-then-Walk (blue) in the same scene. Figures~\ref{fig:baseline_cases_b} and~\ref{fig:baseline_cases_b_part2} isolate the loss of driving efficiency from conservative Walk-then-Run training in AutoDrive-P$^3$~\citep{ye2026autodrive}; Figures~\ref{fig:baseline_cases_a} and~\ref{fig:baseline_cases_a_part2} isolate the collision risk of progress-dominant Walk-and-Run training in ReCogDrive~\citep{li2025recogdrive}.

\textbf{Walk-then-Run (AutoDrive-P$^3$): safe but low progress.}
In Figures~\ref{fig:baseline_cases_b} and~\ref{fig:baseline_cases_b_part2}, the baseline trajectories remain safe but end earlier than ours. This conservative behavior reduces driving efficiency despite avoiding unsafe maneuvers, whereas discovering progress before safety refinement yields longer safe trajectories.

\textbf{Walk-and-Run (ReCogDrive): excessive progress seeking compromises safety.}
In all six examples in Figures~\ref{fig:baseline_cases_a} and~\ref{fig:baseline_cases_a_part2}, the baseline pushes forward but collides: NC is $0$ and the collision-gated PDMS is $0$. Run-then-Walk retains useful forward progress while avoiding the collision (NC $1$ in each case).

\clearpage

\clearpage

\clearpage

\clearpage

\clearpage

\end{document}